**Predicting human decisions with behavioral theories and machine learning**

Ori Plonsky[1], Reut Apel[1], Eyal Ert[2], Moshe Tennenholtz[1], David Bourgin[3], Joshua C. Peterson[4], Daniel Reichman[5], Thomas L. Griffiths[6], Stuart J. Russell[7], Even C. Carter[8], James F. Cavanagh[9], Ido Erev[1]

[1] Technion – Israel Institute of Technology

[2] The Hebrew University of Jerusalem

[3] Adobe Research

[4] Boston University

[5] Worcester Polytechnic Institute

[6] Princeton University

[7] University of California, Berkeley

[8] United States Army Research Laboratory

[9] The University of New Mexico

Correspondence concerning this article should be addressed to Ori Plonsky, Faculty of Data and Decision Sciences, Technion, Haifa, Israel. Email: plonsky@technion.ac.il

**Abstract**

Predicting human decisions under risk and uncertainty remains a fundamental challenge across disciplines. Existing models often struggle even in highly stylized tasks like choice between lotteries. We introduce BEAST Gradient Boosting (BEAST-GB), a hybrid model integrating behavioral theory (BEAST) with machine learning. We first present CPC18, a competition for predicting risky choice, in which BEAST-GB won. Then, using two large datasets, we demonstrate BEAST-GB predicts more accurately than neural networks trained on extensive data and dozens of existing behavioral models. BEAST-GB also generalizes robustly across unseen experimental contexts, surpassing direct empirical generalization, and helps refine and improve the behavioral theory itself. Our analyses highlight the potential of anchoring predictions on behavioral theory even in data-rich settings and even when the theory alone falters. Our results underscore how integrating machine learning with theoretical frameworks, especially those—like BEAST—designed for prediction, can improve our ability to predict and understand human behavior.

Many human decisions in health, finance, environment, and management occur under risk and uncertainty. Understanding and predicting such decisions is a fundamental goal in fields such as economics, psychology, cognitive science, and artificial intelligence. Indeed, decision making under uncertainty has been a central topic of research since Bernoulli's work nearly three centuries ago.[1] Although this research has led to valuable insights and to development of many behavioral models grounded in empirical phenomena and/or theoretical constraints,[2–4] no single model consistently and accurately describes and predicts choices across even the most basic stylized tasks, like choice between lotteries.

Recent large-scale studies have sought to identify a model capable of such robust prediction.[5–7] In one study,[5] a choice prediction competition, researchers submitted models predicting human choice between lotteries, and the models were evaluated based on their predictive accuracy in new held-out data. With the focus on prediction accuracy, one might expect machine learning (ML) tools to excel. Indeed, ML tools have a strong predictive record across domains, including in prediction of human choice under uncertainty,[8–11] and their predictive power is often assumed to provide an upper bound on the possible accuracy of behavioral descriptive models.[12–15] However, the competition and additional analysis have shown that behavioral-theory-free ML performed poorly compared to models incorporating behavioral insights. Chief among these were variants of the behavioral model BEAST (Best Estimate and Sampling Tools).[5] Interestingly, BEAST makes very different assumptions than those assumed by mainstream models like prospect theory. Whereas most models were designed to clarify interesting deviations from expected utility theory, BEAST was designed to predict behavior, and posits that choices result from a potentially biased mental sampling process and sensitivity to expected values.

Subsequent studies revealed boundary conditions on BEAST's dominance. Plonsky et al.[16] demonstrated that an ML algorithm using features derived from the behavioral assumptions of BEAST outperformed BEAST itself—and all other models—on the competition's data. Similarly, Peterson et al.[7] showed that when deep neural networks are designed to reflect theoretical behavioral assumptions, they can efficiently and accurately predict choice in similar tasks. These findings thus suggest that the hybrid approach, combining ML with behavioral features, can harness the strengths of both, augmenting the predictive power of ML with domain-relevant knowledge. However, Peterson et al.[7,17] also demonstrated that with sufficiently large datasets, purely data-driven neural networks can very accurately predict risky choice. This suggests that given enough training samples, behavioral insights may not add much. Consequently, it is unclear which approach best predicts human choice on new data: strictly behavioral models like BEAST, behavioral-theory-free ML trained on ample data, or hybrid models that integrate ML with behavioral theories.

Here, we start by presenting the design and results of another choice prediction competition, CPC18, that expanded the space of choice tasks examined in the original competition and explicitly encouraged submissions that involve ML.[18] A key advantage of the competition format is that it reduces the risk of overlooking alternative modeling strategies by inviting many distinct approaches to the same predictive challenge. The winning submission, from five of the current authors (DB, JCP, DR, TLG, & SJR), was a hybrid model called BEAST-GB (BEAST Gradient Boosting). BEAST-GB combines BEAST's quantitative predictions and features engineered based on the assumptions of BEAST with an Extreme Gradient Boosting (XGB) algorithm.[19] Its success reinforces the idea that combining ML and behavioral logic yields superior predictions of human choice in this domain.

We then proceed to examine the performance of BEAST-GB in two other large datasets, each illuminating different facets of the hybrid approach. First, using the largest public dataset of human choice between lotteries, we test whether BEAST behavioral insights (implemented as features) remain valuable even when the training data is substantially increased. That is, we check if purely data-driven ML can learn the behavioral choice patterns without direct access to domain-specific theoretical logic. We also analyze the differences between the predictions of BEAST and those of BEAST-GB to uncover predictable patterns in choice behavior that BEAST misses. Second, we use a large meta-dataset recently compiled to compare dozens of decision-making models in a different decision-making task to examine how much value ML can add above and beyond the performance of the behavioral model. Specifically, we check whether, even when BEAST itself predicts poorly, a hybrid leveraging its structure still excels. Together, these analyses clarify whether it is truly the integration of BEAST's insights with ML that drives BEAST-GB's success. Finally, we investigate whether BEAST-GB's powerful predictive abilities reflect mere flexibility in capturing idiosyncrasies in each dataset or a broader capacity to capture underlying choice tendencies. We do so by training it on data from some experiments and testing it on different experiments, thereby assessing its context generalization, a pinnacle of predictive modeling.[20]

# Results

## CPC18: A Choice Prediction Competition

Five of the present authors (OP, RA, EE, MT, & IE; hereinafter the organizers) organized CPC18, a choice prediction competition for human choice between lotteries (https://cpc-18.com),[18] a domain that underlies both the foundations of rational economic theory[4,21] and the analyses of robust deviations from rational choice.[2,3] CPC18 used a unified space of decisions under risk, under ambiguity, and from experience (Figure 1), in which at least 14 classical choice anomalies emerge (including St. Petersburg's,[1] Allais'[22], and Ellsberg's[23] paradoxes). Competing participants received choice data from 210 tasks sampled from this space, and were required to predict the distribution of choices in 60 new held-out tasks sampled from the same space (without knowing which tasks would be used for testing during model development). Accuracy was measured by mean squared error (MSE), supplemented by a "completeness" metric which is defined as the fraction of predictable variation in the data that the model captures.[14] It is calculated as $(MSE_{random} - MSE_{model})/(MSE_{random} - MSE_{irreducible})$, where $MSE_{random}$ is the test MSE of a model that assumes random behavior, $MSE_{model}$ is the test MSE of the model, and $MSE_{irreducible}$ is the estimated test MSE of a theoretical perfect model, whose error is only a result of the sampling variation (see methods). With the training data, the organizers also published (before the test data was collected) two baseline benchmarks: A purely behavioral model that is an adaptation of BEAST, and the hybrid model Psychological Forest[16] that uses the behavioral insights of BEAST as features in a random forest[24] algorithm (see Supplemental Information, SI, for details).

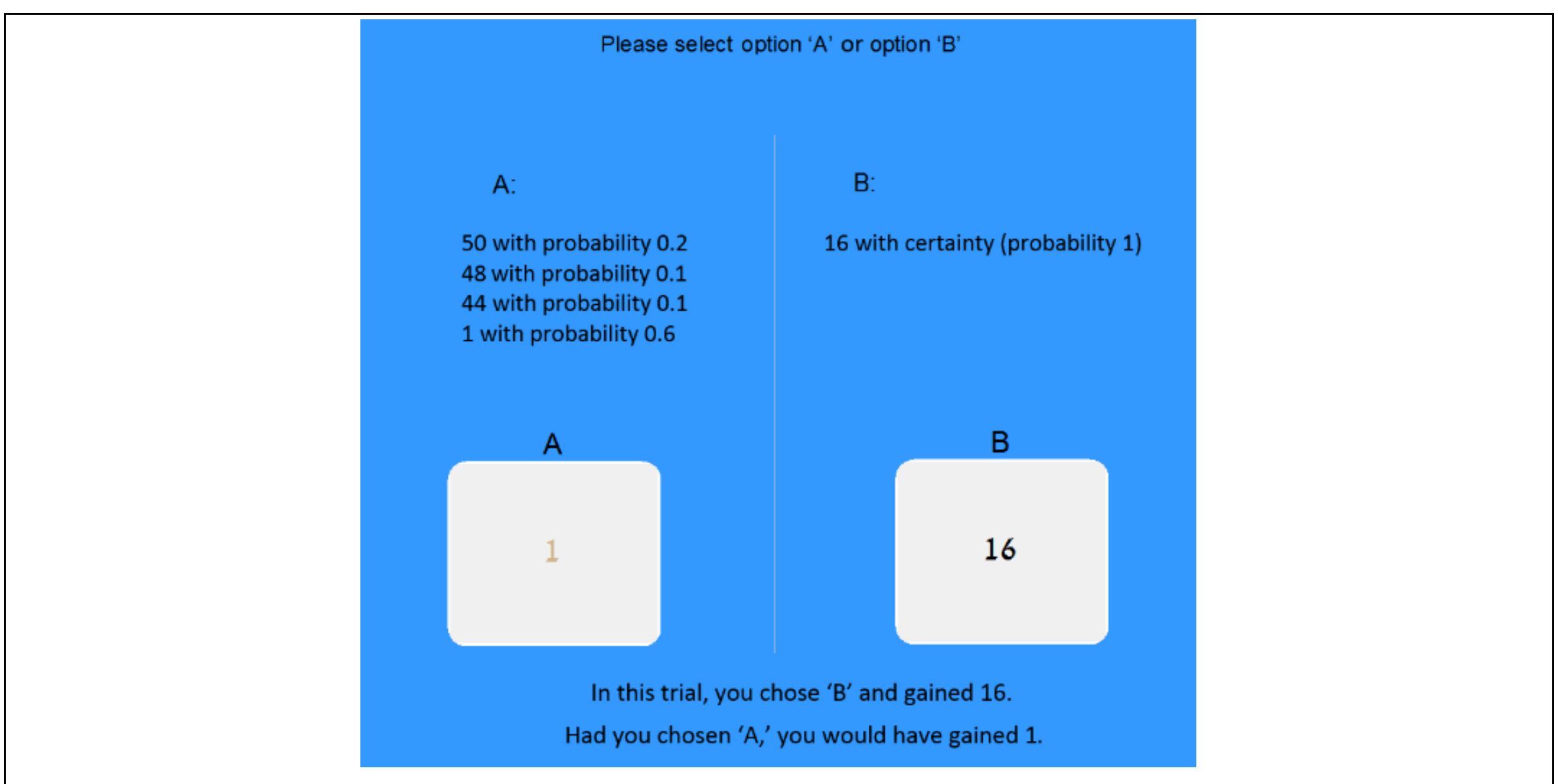

Figure 1. A decision-making task used in CPC18. Human decision makers chose between A and B repeatedly for 25 trials, and got full feedback after each choice starting trial 6. The figure shows the feedback screen after the participant chose Option B.

Forty-six teams, involving 69 researchers representing 34 institutions from 16 countries, registered to the competition. A post-competition survey ($N = 29$; see SI) indicated that many teams invested significant effort; the reported average time spent on model development was 66 hours ($SD = 92$). Twenty models were submitted in time. All submissions integrated behavioral assumptions, suggesting purely data-driven methods struggled in this domain.

The top-ranked submission, BEAST-GB, is an Extreme Gradient Boosting (XGB) algorithm[19] that uses the same features as the baseline hybrid Psychological Forest. BEAST-GB uses, in addition to features describing each task (hereinafter the "objective" features), three sets of "behavioral" features: (1) "Naïve" features that capture naïve intuition for what may matter in choice between lotteries (e.g., the difference between the lotteries' expected values; EVs), (2) "psychological insight" features that were hand-crafted based on the behavioral insights underlying BEAST (e.g., the difference between the probability of each lottery to generate a better outcome, based on BEAST's assumption of simultaneous mental sampling of outcomes from both lotteries), and (3) a "behavioral foresight" feature: the numeric prediction of BEAST itself. Note the distinction between psychological *insight* features, designed to capture a general tendency that can drive behavior, and behavioral *foresight* features, quantitative predictions of behavior in a task (cf. [25,26]). Table S2 details all features used. BEAST-GB achieved 92.6% completeness, capturing nearly all predictable variation in the test data and winning CPC18.

***Analyses of feature importance***

We investigated, using two methods, which features help BEAST-GB most in making such accurate predictions. First, we removed entire feature sets from BEAST-GB, retrained it, and measured the drop in its predictive power. The results (Figure 2a) show that removing the behavioral foresight feature, BEAST's prediction, led to the biggest decline, doubling the MSE and reducing completeness score to 82.8%. This highlights that BEAST alone provides accurate predictions of choice in CPC18. Removing the psychological insight features also degraded accuracy (MSE increased 18%, but completeness remained high). Recall these features were crafted based on the assumptions of BEAST; the fact that removing them hurts performance despite the usage of BEAST itself as a feature implies that they hold information that extends beyond how they are captured in BEAST.

Second, we quantified the feature importances by computing their average absolute SHAP values on the test set. SHAP (SHapley Additive exPlanations), named after the Shapley value in cooperative game theory, is a popular way to compute feature importance in ML.[27] A feature's SHAP value captures its unique contribution to the model's prediction, such that larger absolute SHAP values imply greater importance. Figure 2b shows that the most important feature is the prediction of BEAST, followed by three psychological insight features. These insight features were designed to capture people's assumed sensitivity to the probability that one option provides a better outcome than the other, and to the difference in the best estimates of the EVs.[i] The results of both analyses thus suggest that the behaviorally informed features are vital for BEAST-GB's predictive power.

---

[i] In decisions under ambiguity (which are included in CPC18), direct computation of an option's EV is impossible, and BEAST replaces the EV with its "best estimate". This feature captures this estimate.

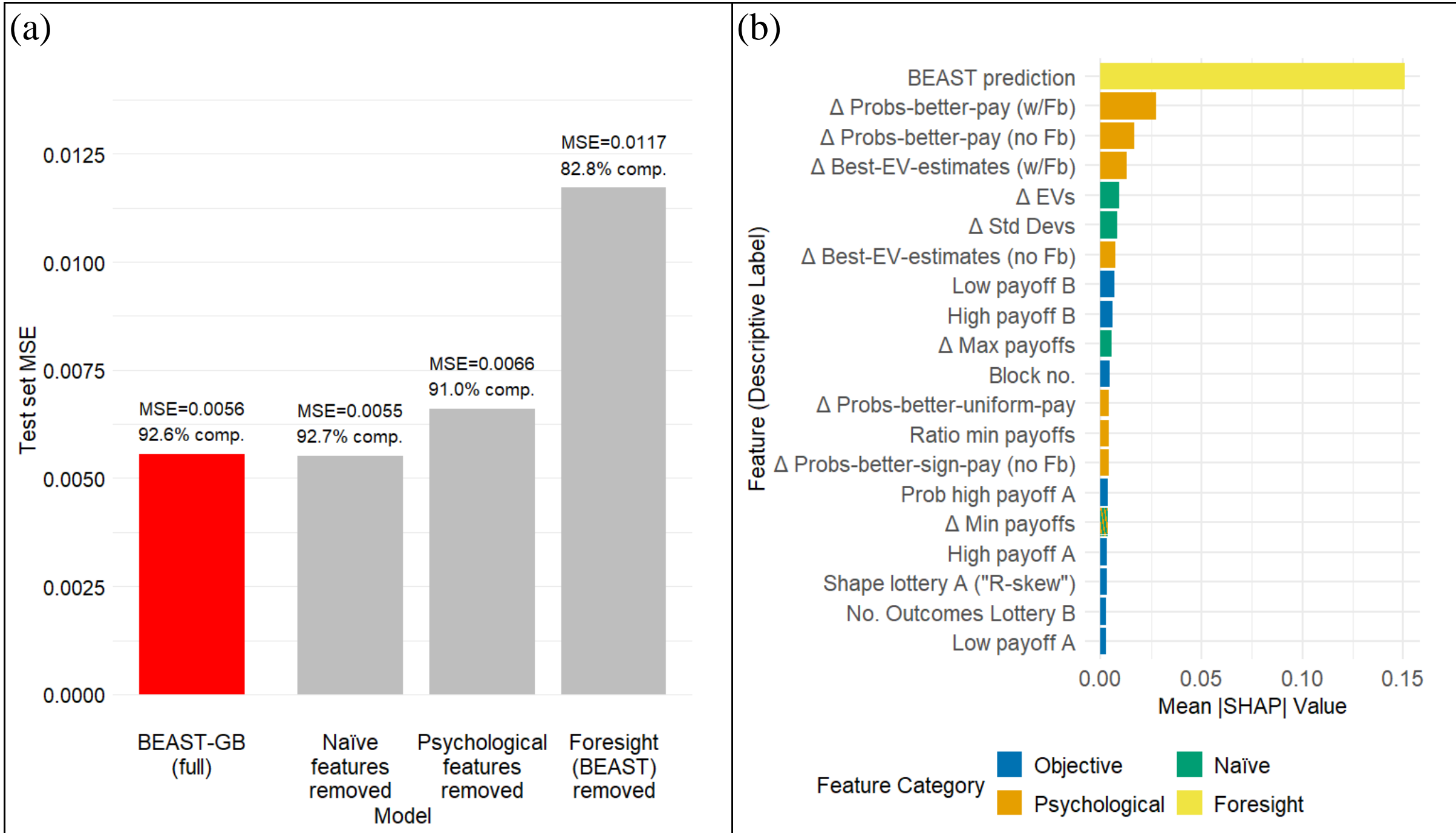

Figure 2. Feature importance analyses for CPC18 data. (a) CPC18 test set predictive performance of BEAST-GB and variations of it that remove different feature sets. (b) Average absolute SHAP values of BEAST-GB's features in predicting CPC18's test set. Only top 20 features are shown. Feature names and definitions appear in Table S2.

***Which "foresight" feature?***

Because the predictions of BEAST were the most useful feature in BEAST-GB, we next tested whether using the predictions of other behavioral models as foresight features would be as useful. We fitted four classical models of decisions under risk, including two variants of Prospect Theory, and used their predictions as features in an XGB trained to predict the competition's data. Notably, we did this for a subset of tasks which was the focus of classical decision research: pure decisions under risk (without ambiguity and without feedback). On this subset, BEAST, which derives its main assumptions from studies of the effects of feedback, should be at a significant disadvantage. Unlike the other models, we also did not specifically fit it to this subset. Nevertheless, Figure S1 shows that BEAST is a far more useful "behavioral foresight" than the other models. Using BEAST as foresight, completeness of the hybrid model is 90%. Replacing it with the second best behavioral model we examined, a version of Cumulative Prospect Theory (CPT),[2] cuts completeness to 67%.[ii] Thus, the BEAST-derived foresight signal proved uniquely powerful.

---

[ii] We also examined each model's standalone performance (without feeding its predictions into XGB). We found that, in every case, using the model's predictions as a feature in XGB performed better. Furthermore, including all five behavioral "foresights" in XGB did not outperform using BEAST as the sole foresight feature.

**Choices13k: Behavioral theory when data is abundant**

The results of CPC18 highlight BEAST's usefulness in predicting human choice between lotteries and demonstrate that integrating its predictions and behavioral insights into a ML algorithm yields further gains. In many real-world prediction problems, training data is rather limited, for example because when implementing a new incentivization policy, only few treatments can be piloted before choosing a policy. However, the training data in CPC18 is considerably smaller than in many tasks in which ML algorithms that are not ingrained with theoretical domain knowledge excel. This raises the question of whether behavioral theory remains necessary when the training data is large. It is possible that with more data, purely data driven methods can learn the regularities captured by BEAST (and/or other theories) directly, so behavioral insights only matter when data is scarce.[17]

To explore this, we evaluated BEAST-GB on Choices13k, the largest publicly available dataset of human choice under risk and uncertainty.[17] It includes nearly 10,000 choice tasks similar to those used in CPC18 (see Methods and Table S5 for main differences). Importantly, using Choices13k, prior studies have shown that with such large data, ML algorithms—specifically deep neural networks—could achieve high accuracy even without built-in behavioral logic (though training was more efficient with it).[7,17] Following Peterson et al.,[7] we repeatedly split the dataset into training (90%) and test (10%) sets, and trained models on increasingly larger fractions of the training data. This procedure allows checking how much data is needed to reach different levels of predictive accuracy.

Figure 3 compares BEAST-GB with several benchmarks, including Context-Dependent (CD), one of the best and most expressive neural networks analyzed in Peterson et al. BEAST-GB achieved state-of-the-art accuracy (MSE = 0.00843), with 96.2% completeness, capturing nearly all predictable variation in the data. Furthermore, BEAST-GB required relatively few training examples to reach high accuracy: Trained on just 2% of the training data (176 choice tasks), it already predicted more accurately (MSE = 0.0110) than CD trained on all ~9000 tasks (MSE = 0.0113). This highlights how incorporating behavioral logic can dramatically improve sample efficiency, enabling models to achieve strong predictive performance with substantially less data.

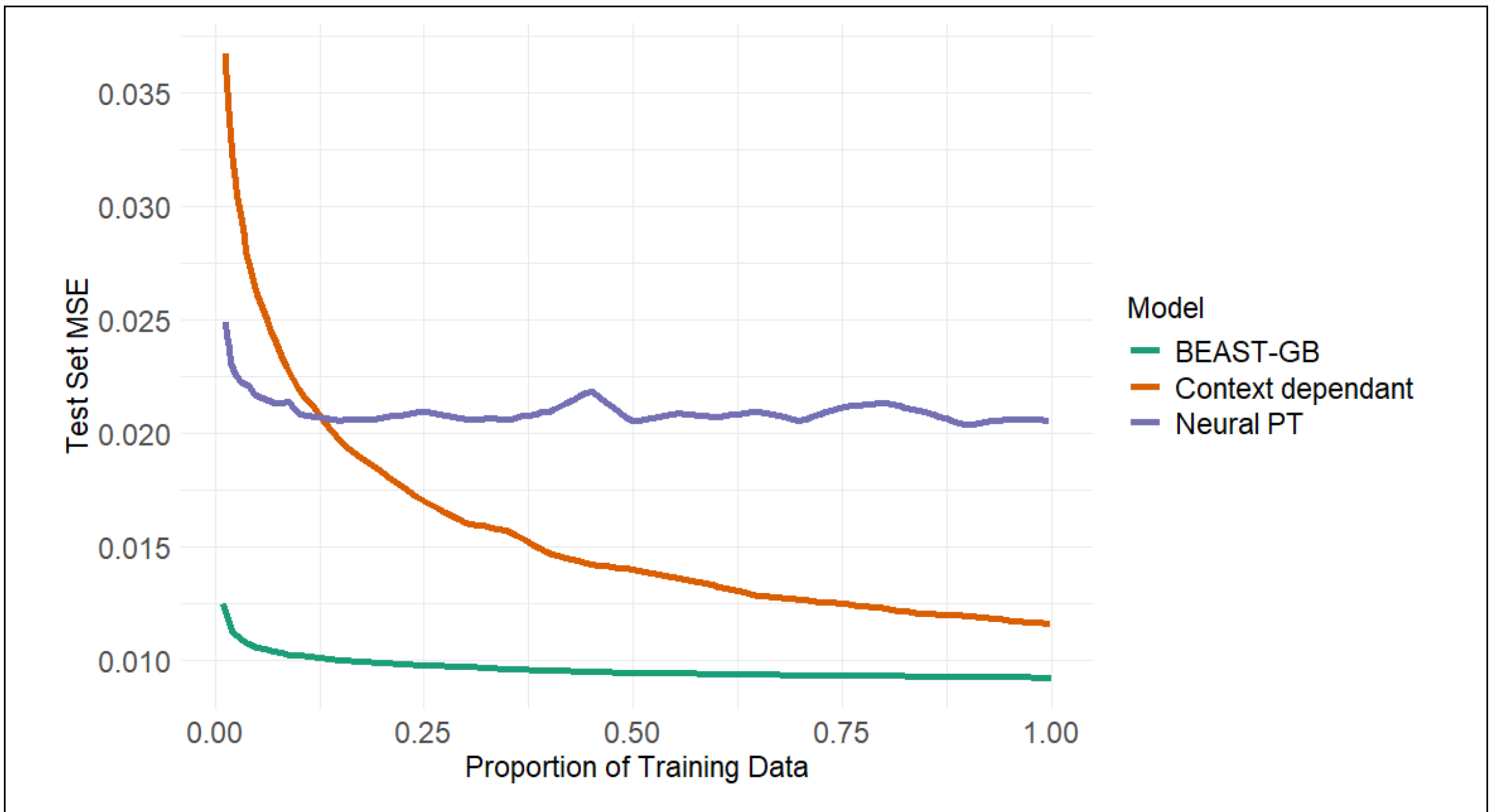

Figure 3. Test set performance on Choices13k data. Data was split to 90% training and 10% held-out test data, and models were trained on fixed and increasing proportions of the training data. This process was repeated 50 times, and results reflect the average test set MSE over these 50 splits. Neural PT (Neural Prospect Theory) and Context dependent performance is taken directly from Peterson et al. (2021).

Analyses of feature importance confirmed that behavioral features remain critical, even in this data-rich environment, though the influence of BEAST's predictions as a foresight feature diminishes with increasing data availability. Figure S2a shows that when training data was scarce, removing BEAST's prediction feature sharply impaired performance, suggesting that BEAST provides an effective initial approximation that helps mitigate bias (see SI for Bias-variance analyses). Yet, with sufficient data, the removal of the foresight feature was inconsequential (MSE = 0.00853, not significantly different than BEAST-GB, $t(49) = -1.24$, $p = .22$, ΔMSE = −0.0001, 95%CI = [−0.0003, 0.0001], $BF_{10}$ = 0.32). This suggests that, as training data increases, the model can learn a proper integration of the psychological insights underlying BEAST without direct access to BEAST itself. In contrast, removing psychological insight features—hand-crafted to reflect BEAST's behavioral mechanisms—reduced accuracy even with the full dataset (MSE = 0.00879; significantly worse than BEAST-GB, t(49) = −5.09, p < .001, ΔMSE = −0.0004, 95%CI = [−0.0005, −0.0002]). Further, removing both the psychological insight and foresight features worsened performance still (MSE = 0.00920), and using only objective task features drastically reduced accuracy (MSE = 0.01530). Thus, even when data was abundant, purely data-driven models failed to fully capture the predictive power of behavioral insights. Analysis of SHAP values (Figure S2b) further supported these conclusions.

Interestingly, models trained using only behavioral features, without access to objective task structure, also performed significantly worse than BEAST-GB (MSE = 0.00914; $t(49) = -8.08$, $p < .001$, $\Delta MSE = -0.0007$, 95%CI = [−0.0009, −0.0005]). This suggests that while task structure alone carries little predictive power, it provides crucial context for behavioral features to be effectively leveraged. These results imply that some of BEAST-GB's success stems from an integration of task structure and BEAST's behavioral logic.

***Using BEAST-GB to explain behavior***

If the predictive power of BEAST-GB involves successful integration of task structure and the behavioral insights of BEAST, it should be possible to identify classes of tasks in which BEAST-GB predicts systematically differently than BEAST. Because BEAST-GB captures nearly all of the predictable variation in the data, analyzing where it deviates from BEAST can reveal choice patterns that BEAST overlooks, potentially informing improvements to the behavioral model itself. Note that this process is more effective than critiquing BEAST with respect to the data because its deviations from the observed behavior also reflect unpredictable noise.[15]

Our analysis of the differences between BEAST's and BEAST-GB's predictions showed that 90% of the variance in the deviations could be explained by three sets of intuitive corrections (see SI). First, BEAST's predictions are too extreme, especially when task complexity increases, implying BEAST-GB identifies that behavior in the (online) experiments of Choices13k is noisier than BEAST (trained on lab data) expects.[28] Second, BEAST fails to capture systematic "gain seeking" tendency in tasks that involve the possibility to avoid a sure loss. This behavior contradicts loss aversion but is consistent with the experimental design in Choices13k, where negative payments were replaced with zero. Third, BEAST assumes that each of its mechanisms operates uniformly across all tasks, but BEAST-GB can dynamically adjust their relative influence based on task structure. Some of the systematic deviations of BEAST from BEAST-GB, like gain-seeking in Choices13k, are likely dataset-specific, but others can be more general. Indeed, insights from this analysis led to a simple correction to BEAST, which—without increasing complexity or reducing interpretability—improved its predictive performance across all datasets considered in this study (see SI).

These findings demonstrate how hybrid models like BEAST-GB not only enhance prediction accuracy but also serve as a powerful tool for refining and improving behavioral theories. By leveraging the flexibility of ML while preserving interpretability, BEAST-GB reveals systematic choice patterns that would otherwise be obscured by theoretical constraints or data noise.

**HAB22: Machine learning when theory fails**

Although BEAST-GB achieved high accuracy in both CPC18 and Choices13k, these successes may have arisen primarily because BEAST itself is already very effective for those datasets. Indeed, in CPC18, the second-best submission was a minor modification of BEAST (Table S1), and a scalable retrained variation of BEAST performed similarly to CD with completeness of 88.3%.[29] Furthermore, the original BEAST, without retraining of its parameters, already captured much of the predictable variation in both CPC18 (88.9% completeness) and Choices13k (65.7%). Thus, it remains unclear how much real benefit comes from merging a strong behavioral model (BEAST) with ML, as opposed to simply relying on the behavioral model alone.

To investigate this question, we turned to a dataset of risky-choice tasks recently collected by He, Analytis, and Bhatia, henceforth, the HAB22 dataset.[6] HAB22 differs from CPC18 and Choices13k in several important ways (see Figure S3 and Table S5). First, it includes data from multiple distinct experimental contexts. Second, it is restricted to choice between lotteries with up to two outcomes and without feedback. Last, it includes data from experiments designed to produce strong context effects.[30] In many of the tasks from these experiments, the lotteries' EVs differed dramatically, and—potentially because of context effects—participants often did not choose the option with the much higher EV.[31,32] While such tasks are useful for demonstrating interesting deviations from expected utility theory, using them can hurt BEAST's predictive power, as it assumes high sensitivity to EV differences. Thus, HAB22 allowed us to examine the generality of our results in several ways, as we show below.

Originally, HAB22 was used to evaluate 53 existing behavioral models by fitting them to each participant's data and then predicting the same individuals' choices on new tasks. BEAST-GB (like BEAST) was designed for predicting behavior of new decision makers in new tasks, not for predicting known individuals. In the SI, we demonstrate that models relying on BEAST-GB's population-level predictions, together with BEAST's underlying logic, predict the individual choices in HAB22 as well as or better than the best extant behavioral models. For consistency with our other analyses, however, here we compared BEAST-GB to the behavioral models in predicting aggregate choice rates for new participants facing new tasks.[iii]

This comparison revealed that on HAB22, the original BEAST (without retraining) fared poorly, achieving only 36% completeness. In contrast, the strongest purely behavioral model, a version of CPT, reached 93.8% completeness (MSE = 0.0316). Nevertheless, as shown in Figure 4, BEAST-GB, which used BEAST's very inaccurate predictions as a feature, outperformed all other models, with completeness of 94.8% (MSE = 0.0307). This improvement over the best behavioral model was significant, $t(49) = 2.99$, $p = .004$, ΔMSE = 0.0010, 95%CI = [0.0003, 0.0016].

Interestingly, removal of the "foresight" feature hurt the model's performance (MSE = 0.0313, significantly worse than BEAST-GB, $t(49) = -4.08$, $p < .001$, ΔMSE = −0.0006, 95%CI = [−0.0009, −0.0003]), and this feature remained the most important according to SHAP value analysis (Figure S4). Hence, BEAST holds important information concerning behavior even when its raw predictions are poor. One reason for this is likely the high rank-order (Spearman) correlation ($\rho = 0.819$) between (the untrained) BEAST's predictions and the observed choice rates. Additionally, we show in the SI that most of the differences between the predictions of BEAST and BEAST-GB are accounted for when the mechanisms in BEAST are rescaled for each experimental context and task structure. This implies that the ML component in BEAST-GB identifies and corrects BEAST's context-dependent miscalibrations, enabling superior accuracy even when BEAST alone performs poorly.

[iii] For robustness, we also compared BEAST-GB to the behavioral models when the training and test data use the same participants (facing new tasks). Although the behavioral models were individually fitted to the participants whose behavior they should predict (and BEAST-GB was not), the results show that BEAST-GB still predicts the aggregate behavior significantly better than all behavioral models (see SI).

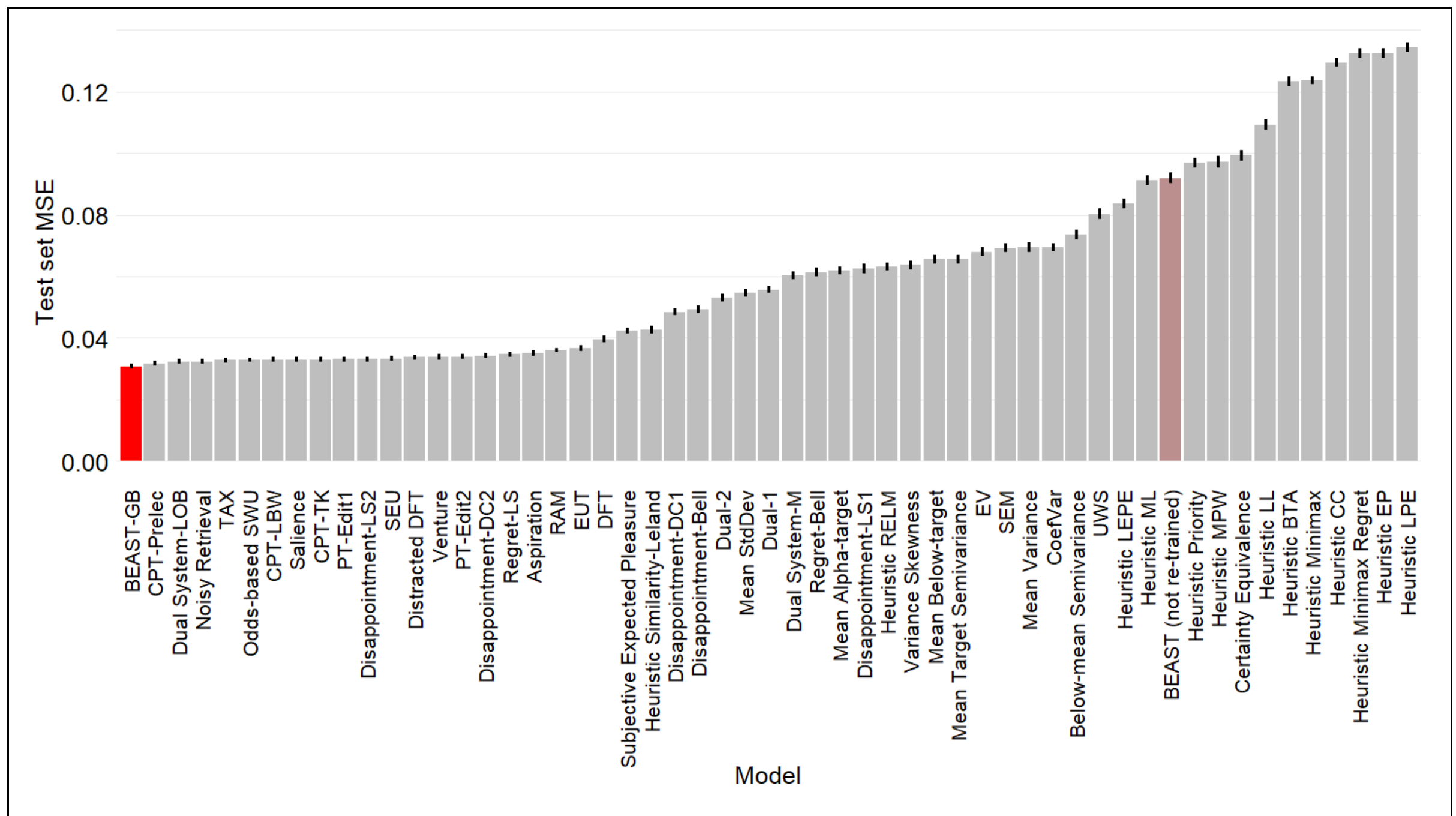

Figure 4. Test set performance on HAB22 data. Performance is evaluated based on 10-fold cross validation on choice tasks, and 5-fold cross validation on participants in experimental contexts. That is, models predict choice rates of new participants in new tasks (see Methods). Error bars represent ±1 SE for the mean over the 50 test sets. Model names and sources in Table S6.

### Context Generalization

The preceding analyses showed that, for each of three large datasets of human choice under risk and uncertainty, training BEAST-GB on tasks from within the same context yielded highly accurate predictions of new tasks. We next asked whether BEAST-GB, when trained on choice data from one experimental context, could effectively predict behavior in a different experimental context. The ability to generalize across contexts—often called domain or context generalization—is a highly desirable property of predictive models.[20,33,34] Furthermore, recent work suggests that different experimental contexts in decisions under risk and uncertainty can systematically differ in subtle but important ways, meaning a model trained on one context can struggle when tested on another.[28]

To examine BEAST-GB's capacity for context generalization, we exploited the fact that HAB22 is a collection of distinct experimental contexts. We systematically trained BEAST-GB on all but one of the contexts, then predicted behavior in the held-out context, without using its choice tasks or participants during training. On average, BEAST-GB yielded MSE of 0.0162 in the unseen context, corresponding to 87.2% completeness (SD =

0.08). That is, without access to data from the target context, BEAST-GB achieved over 87% of the predictive accuracy expected from a perfect hypothetical model that knows the population parameter for each task in that context.

Furthermore, 31% of the choice tasks in HAB22 appeared in more than one experimental context. This allowed us to compare BEAST-GB's generalization capacity (i.e., its accuracy in predicting behavior outside of context) to direct empirical generalizations, namely, to using the *observed* choice rate of a given task in one context as a prediction to the choice rate of the *same task* in another context. Note that since people in different contexts do not necessarily behave similarly and given sampling errors, quantitative models trained to capture general patterns of behavior across tasks might predict more accurately the choice rate in the new context. That is, the error of the predictive models could potentially be smaller than the average sampling error. As Figure 5 shows, none of the behavioral models examined by He et al. achieved this feat, but BEAST-GB did. Its MSE when predicting choice rates of known tasks in new experimental contexts was 0.0121 (91.8% completeness), representing a 13% improvement over simply assuming behavior directly generalizes across experimental contexts and predicting the same task's observed choice rate from the training contexts. The difference is significant, $t(827) = -3.79$, $p < .001$, $\Delta MSE = -0.0019$, 95%CI = [−0.0028, −0.0009]. That BEAST-GB usefully predicts choice behavior in new contexts it was not trained on suggests that it captures generalizable choice tendencies, rather than merely fitting idiosyncratic patterns from specific samples of tasks and participants.[34]

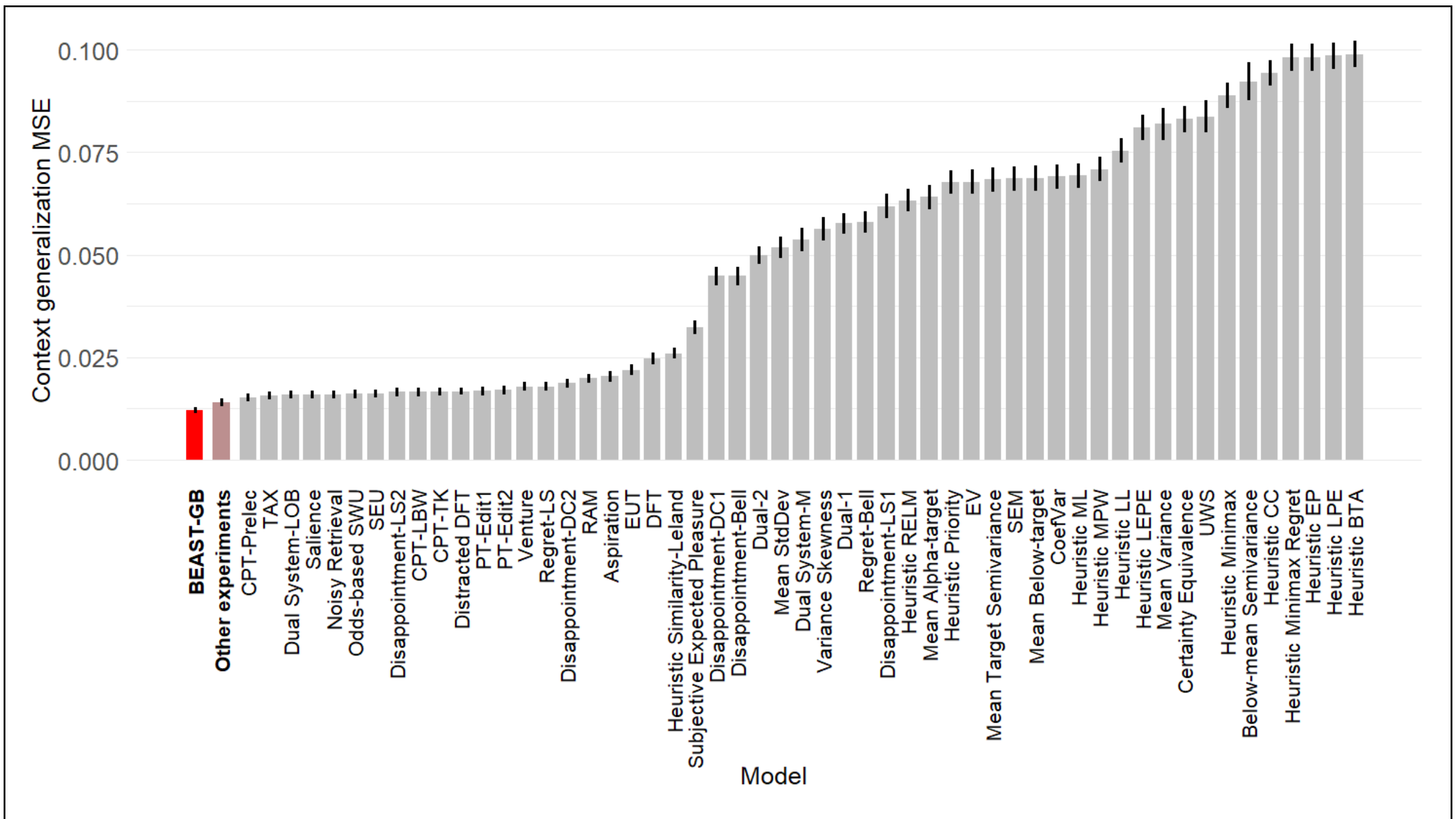

Figure 5. Predictive accuracy in context generalization task of predicting behavior in the 828 instances where a choice task that appears in the test dataset also appeared in one or more of the train datasets. Training data always includes 15 experimental contexts to predict the 16$^{th}$ context. Behavioral models' predictions are set to be the average training prediction (i.e., best fit) in the target task across all subjects in the training data. "Other experiments" prediction is the average observed behavior across all subjects in the training data in the target task. Error bars represent ±1 SE for the mean over the 828 prediction errors. Model names and sources in Table S6.

## Discussion

Our paper introduces BEAST-GB, a hybrid model that integrates a strong behavioral theory (BEAST) with ML to predict human choice under risk and uncertainty. Across three datasets encompassing more than 11,000 choice tasks, BEAST-GB demonstrated state-of-the-art predictive accuracy, consistently capturing over 92% of the predictable variation that would have been captured by a perfect (hypothetical) model. BEAST-GB won an open prediction competition featuring genuinely independent test data,[35,36] and maintained predictive superiority within the largest public dataset of risky choice as well as a collection of 15 distinct experimental contexts that differ in participant pools, settings, and methodologies. Furthermore, BEAST-GB successfully generalized across contexts, outperforming even direct empirical generalizations from observed behavior. These findings underscore BEAST-GB's broad applicability across decision-making environments.

Our analyses suggest that BEAST-GB's predictive success stems from the effective synergy between behavioral theory and ML. The integration involves three categories of features: objective task characteristics, psychological insights that represent BEAST's behavioral mechanisms, and foresight provided by BEAST's quantitative predictions. The foresight feature supplies the ML algorithm with an initial powerful signal about likely behavioral patterns. The ML component then adjusts these predictions by dynamically weighting the psychological insights underlying BEAST across diverse task structures. Fully disentangling these adjustments is difficult, so using our hybrid approach purely to explain the cognitive processes behind choice behavior remains limited. Nevertheless, we show that some of the adjustments can be explicitly identified, offering insights into systematic limitations within BEAST itself. Notably, these insights led to a refinement of the original behavioral model, highlighting that such hybrid models can also serve as theoretical diagnostic tools capable of improving our understanding of decision-making processes.

A notable advantage of the hybrid approach is that it provides an efficient way to scale rigid and complex behavioral models like BEAST to new and large datasets without extensive re-fitting. BEAST, by itself, involves computationally demanding simulations that make training on new data cumbersome. BEAST-GB circumvents these limitations, rapidly adjusting BEAST's theoretical predictions and insights to new contexts. Furthermore, because BEAST was developed to capture behavior across a wide set of situations, it includes rigid constraints that are not necessarily theoretically grounded but help it avoid overfitting. Some of these constraints, however, may limit BEAST's adaptability. For example, BEAST gives high weight to the best estimates of the EVs, contributing to its lower predictive accuracy in some contexts (specifically some experimental contexts in HAB22 that showcase low sensitivity to EVs).[37] BEAST-GB utilizes the information embedded in BEAST while effectively avoiding this bias. This demonstrates that the scalability afforded by hybrid models can advance behavioral research by enabling exploration of complex phenomena without the constraints imposed by the behavioral models' architectural rigidity.

Another strength of our hybrid approach is its generalizability. Indeed, the approach underlying BEAST-GB is not restricted to predicting choices between lotteries. In the SI, we demonstrate that a similar hybrid approach can achieve state-of-the-art predictive accuracy in an entirely different decision domain, two-player extensive form games. For this approach to be effective in other domains, the key requirement is a foundational behavioral model that is reasonably accurate and sufficiently broad to provide meaningful behavioral insights that ML can exploit and extend. Hence our hybrid approach is naturally limited by the theoretical basis and generalization capacity of extant models in each behavioral domain.

**Implications for research in behavioral science**

Behavioral science predominantly seeks explanations of behavior, leading many researchers to focus on discovering new phenomena and specifying causal mechanisms.[38] Although such work is invaluable, it inevitably leads to the study of narrow scenarios designed to illustrate specific phenomena. For example, behavioral decision-making research focuses on situations that demonstrate deviations from rational choice. Moving from elegant but narrow explanations toward robust and useful predictions—essential also for validating underlying theoretical mechanisms— requires greater emphasis on identifying behavioral principles that reliably generalize across broader sets of tasks.

We speculate this rationale explains why BEAST, and its underlying mechanisms, provide highly useful behavioral insights for BEAST-GB. Unlike classical models primarily designed to capture anomalies where the rational benchmark is obvious, BEAST was originally developed to predict choice across a broad set of situations, including decisions under ambiguity and from experience. Its main assumptions are grounded in fundamental learning processes (e.g., that choice is sensitive to the probability of obtaining better outcomes), many of which are shared across species,[39,40] highlighting their potential generality and robustness. By considering a broad spectrum of situations, BEAST's developers could identify generalizable and useful insights that proved critical in enhancing BEAST-GB's predictive robustness and applicability.

**Conclusion**

Our research advances behavioral decision-making research by demonstrating the power of hybrid models that integrate behavioral logic with ML. BEAST-GB's success across diverse datasets and tasks and its ability to generalize across experimental contexts sets a new benchmark for accuracy and generalizability in the field. Looking forward, the integration of theoretical insights derived from a prediction-focused approach to behavioral science[41] with ML offers a promising avenue for developing more adaptable, accurate, and generalizable models of human behavior.

# Methods

## Model Evaluation

Throughout the paper, we evaluated models using their Mean Squared Error (MSE) between the predicted and the observed choice rates across tasks in the (test) data. The MSE was recently recommended as the preferred measure for evaluation of behavioral models as it satisfies all desired properties of a loss function in this domain.[42] In addition, to help interpret the accuracy of the models, and following the suggestion of Fudenberg et al.,[14] we computed the models' completeness score, measured as the proportion of predictable variation in the data that the model captures. Completeness equals $(MSE_{random} - MSE_{model})/(MSE_{random} - MSE_{irreducible})$, with $MSE_{random}$ the MSE of random guessing (as defined in Fudenberg et al.[14]), $MSE_{model}$ is the MSE of the model in question, and $MSE_{irreducible}$ is an irreducible error, that is the portion of the total error considered unpredictable. To get $MSE_{irreducible}$, we aimed to estimate the expected MSE of a perfect hypothetical model that accurately predicts the *population* choice rate in a task. Notably, the computed MSE of such perfect theoretical model would likely be positive since models are evaluated based on their accuracy in predicting *estimates* of the population choice rates, namely the observed sample choice rates. That is, the observed error of a perfect theoretical model in task *i* is the sampling error, and thus the computed MSE of this model is equal to the average (over choice tasks) of the squared sampling errors. Since the expectation of the squared sampling error equals the variance of the sample average, we get:

$$MSE_{irreducible} = \frac{1}{N}\sum_{i=1}^{N}\left(\widehat{\mu_i} - \overline{x_i}\right)^2 = \frac{1}{N}\sum_{i=1}^{N}\left(\mu_i - \overline{x_i}\right)^2$$

$$E\left(MSE_{irreducible}\right) = \frac{1}{N}\sum_{i=1}^{N} E_i\left(\mu_i - \overline{x_i}\right)^2 = \frac{1}{N}\sum_{i=1}^{N} Var\left(\overline{x_i}\right) \cong \frac{1}{N}\sum_{i=1}^{N}\frac{S_i^2}{n_i}$$

Where $\widehat{\mu_i}$ is the prediction of the perfect hypothetical model for task *i*, $\overline{x_i}$ is the observed choice rate in the sample for task *i*, $\mu_i$ is the true population choice rate, $S_i^2$ is the sample variance of task *i*, $n_i$ is the sample size for task *i*, and *N* is the number of choice tasks. That is, we estimated $MSE_{irreducible}$ as the average of the squared standard errors.

## BEAST-GB model

BEAST-GB is an XGB algorithm that uses the features detailed in Table S2. Most features used by BEAST-GB are derived from the behavioral model BEAST (Best Estimate and Sampling Tools) designed to predict human decision-making under risk and uncertainty at the population level.[5]

Theoretically, BEAST is grounded in the idea that people adapt strategies that proved effective in past situations perceived as similar to the current one.[43–45] It assumes individuals act as intuitive classifiers: a current task is classified alongside similar previous ones, and a strategy that worked well in that class is invoked.[46] Because the classification can be imperfect, the chosen strategy may sometimes be ill-suited to the current context, resulting in behavioral "biases." Instead of explicitly modeling this complex, individual, and idiosyncratic classification process, BEAST approximates its main implications for the aggregate behavior in risky and uncertain decisions by assuming people in these contexts primarily rely on five cognitive strategies. These strategies are choosing options that (a) are best in expectation, (b) minimize immediate regret, (c) maximize the chances to get a better payoff sign, (d) maximize the worst possible payoff, and/or (e) yields a better payoff if all outcomes were equally likely. The output of the first strategy is computed explicitly or based on one's "best estimate" of the EV (if direct computation is impossible). The output of the other four strategies is implemented via a mental sampling process involving potentially biased "sampling tools" (see SI for the implementation details). Each of the five cognitive strategies was previously translated into psychological insight features,[16] which are now used in BEAST-GB.

XGB (Extreme Gradient Boosting)[19] is an algorithm that efficiently and effectively implements the idea of Gradient Boosting. Gradient Boosting is an iterative ensemble procedure in which simple regression trees—models that repeatedly split the data based on threshold conditions, thereby creating piecewise-constant predictions—are added one at a time to reduce the errors of the existing model. Each new tree learns to predict the residuals from the previous round, so that, over many iterations, the ensemble flexibly models nonlinearities and interactions among features, in a context-dependent manner. To reduce overfitting and improve generalization, XGB includes additional regularization, as well as random selection of features to be used in each iteration. In BEAST-GB, the algorithm takes as input both the objective features that capture the structure of the task and behavioral features that capture behaviorally relevant properties of the task. The algorithm then iteratively learns when and how to utilize them, searching at each iteration for feature interactions that best reduce the remaining prediction errors. The result is an ensemble that effectively ties together the signals available in the various features.

We implemented the following pipeline to train BEAST-GB on each choice dataset. First, we generated the features for each choice task. This notably includes generating the choice rate prediction of the original BEAST model for that choice task. Note that BEAST is not refitted to the new data. Its predictions (to be used as foresight feature) are derived using the original values of parameters fitted to the training set of CPC15 (see SI).[5] Second, we coded categorical features to numeric using dummy coding. Third, because in particular datasets some features may turn out completely constant and/or duplicates of other features, we removed such features from the data. Fourth, we randomly split the data to a train and a held-out test set (unless the data was already organically split, like in CPC18). Fifth, we standardized all features by subtracting their average and dividing by their standard deviation in the train set. Sixth, we tuned the algorithm's hyperparameters using five repetitions of 5-fold cross validation implemented on the train set (see Table S4 for the values of the hyperparameters in each dataset). Finally, we trained the algorithm on the full train set with the chosen hyperparameters and generated its predictions for the held-out test set.

## Feature importance analyses

Throughout the paper, we assessed the relative importance of features included in BEAST-GB for prediction using two distinct methods. The first involved systematically removing sets of features from the tuned model, retraining it on the train set, and evaluating the predictions of the new model (i.e., without the removed features) on the test set. The second method involved computing the mean absolute SHAP values (using package SHAPforxgboost[47] in R) over all predictions of the test set (or, when models were evaluated using multiple iterations using different test sets, all predictions of the test sets).

## CPC18

### *Experimental task*

Similar to the paradigm used in CPC15,[5] the experimental paradigm in CPC18 involved binary choice under risk, under ambiguity, and from experience. As seen in Figure 1, decision-makers were presented with two lotteries (Option A and Option B) and were asked to choose between them repeatedly for 25 trials. In the first five trials, they did not get any feedback, but starting from the 6$^{th}$ trial, they received full feedback concerning the outcomes generated by each option (both the obtained and the forgone payoffs were revealed). Choice options in CPC18 may include up to 10 outcomes, may involve ambiguity (i.e., probabilities of potential outcomes of one of the options were not revealed to the decision maker), and may be correlated between them. A choice task is thus uniquely defined by 12 dimensions: five determine the outcome distribution of Option A ($L_A$, $H_A$, $pH_A$, $LotNum_A$, $LotShape_A$), five determine the outcome distribution of Option B ($L_B$, $H_B$, $pH_B$, $LotNum_B$, $LotShape_B$), one (*Amb*) determines if the task involves ambiguity, and one (*Corr*) determines whether the outcomes in the two options are correlated. See the SI for more details on these dimensions and how they define the tasks.

The space of choice tasks that is implied by these dimensions extends the space studied in CPC15 by allowing both options (rather than just one) to have up to 10 outcomes. Within this space, it is possible to replicate 14 classical behavioral decision making phenomena:[5] the Allais' paradox,[22], the reflection effect,[3] overweighting of rare events,[3] loss aversion,[3] St. Petersburg's paradox,[1] Ellsberg's paradox,[23] low magnitudes eliminate loss aversion,[48] the break-even effect,[49] the get-something effect,[50] the splitting effect,[51] underweighting of rare events,[52] the reversed reflection effect,[52] the payoff variability effect,[53] and the correlation effect.[54]

***Experimental data***

The data used in CPC18 includes 694,500 decisions made by 926 different decision-makers across 270 binary choice tasks. Tasks were divided into 9 cohorts. Each decision-maker faced one cohort of 30 tasks in random order and made 25 choices in each task. The first five cohorts were also used in CPC15,[5] and details on these data are provided elsewhere. The choice tasks in the four additional cohorts were randomly selected from the space of tasks investigated in CPC18 according to a pre-defined task selection algorithm (see SI). Two cohorts of choice tasks were then run in each of two new experiments that used the same participant pool and a very similar design to those used for CPC15.

Each experiment involved 240 participants (Experiment 1: 139 females, $M_{\text{Age}} = 24.5$, $\text{Range}_{\text{Age}} = [18,37]$; Experiment 2: 141 females, $M_{\text{Age}} = 24.7$, $\text{Range}_{\text{Age}} = [18,50]$), mostly undergraduate students, participating in one of two (physical) lab locations: the Technion and the Hebrew University of Jerusalem. No statistical methods were used to pre-determine sample sizes but our sample sizes are larger than those used in previous publications focusing on predictions of choice under risk and uncertainty.[5,6,17] Informed consent was elicited from all participants at the beginning of the experimental session. The experiment lasted approximately 45 minutes. Participants were paid for one randomly selected choice they made, in addition to a show-up fee. The final payment ranged from 10 to 136 shekels, with a mean of 40 (about 11 USD) for Experiment 1 and from 10 to 183 shekels, with a mean of 41.9 for Experiment 2. The experiments complied with all ethical regulations and were approved by the Social and Behavioral Sciences Institutional Review Board in the Technion and by the Ethics Committee for Human Studies at the Faculty of Agriculture, Food, and Environment at the Hebrew University of Jerusalem.

***Competition procedures and protocol***

In May-June 2017, the organizers ran Experiment 1. They then used the combined data from Experiment 1 and from CPC15 to develop their baseline models (see SI) and made the data publicly available. In January 2018, they published a call to participate in the competition in major mailing lists and on social media. The competition included two independent challenges, and in this paper, we focus on the first (see SI for details on the second). In that challenge (or track), the goal was to provide, for each of 60 choice tasks from Experiment 2, run in June-July 2018, a prediction for the progression over time of the mean aggregate choice rate of one of the options. Specifically, the 25 trials of each task were pooled to five blocks of five trials each, and the goal was to predict the mean aggregate choice rates of Option B in each of the five time-blocks. Since the exact nature of the tasks was unknown to modelers at the time of model development, a competing model was required to get as input the values of the 12 dimensions defining each task and provide as output a sequence of five predictions (each in the range [0,1]) for the mean choice rates in that task.

Interested participants were required to register for the competition in advance. Each person could register as a (co-)author of no more than two submissions per track and be the first author of no more than one submission per track. In addition, each person could make one additional early-bird submission, sent to the organizers by the end of January 2018. Submissions had to be made on or before the Submission Deadline (July 24th, 2018). In practice, this meant sending the organizers a complete, functional, documented code of the submission. The code could have been written in Python, R, MATLAB, or SAS. The code was required to read the dimensions of a choice task and provide as output a prediction for the choice rates in the five blocks. One day after the Submission Deadline, the organizers published the test set tasks (the 60 tasks from Experiment 2). That is, submissions were blind to the tasks on which they were tested. Participants then ran their code on the test set tasks and submitted the predictions. Finally, the organizers published the data to be predicted so participants could evaluate their prediction error. The organizers verified that the code for each of the top 10 submissions produces the reported predictions and published the results.

***Statistical significance***

Because ranking of submitted models may depend on the (random) selection of the competition's test set tasks, we used a bootstrap analysis (using Package boot[55] in R) to compare each submitted model with the competition's winner BEAST-GB. Specifically, we simulated 10,000 sets of 60 test choice tasks each by sampling with replacement from the original test set, computed the MSE of each submission in each simulated set, and then counted the number of sets in which a submitted model outperformed BEAST-GB. The proportion of test sets in which a model would have outperformed the winner is the estimated *p*-value for the difference between the winner and the model: If it is smaller than .05, then BEAST-GB is considered to predict significantly better.

***Foresight comparison analysis***

To compare the value of using BEAST as a foresight feature with the value of using other classical decision models as foresight features, we used a subset of the CPC18 data which includes only decisions under risk without feedback: choices made in tasks without ambiguity in the first block of five trials in each choice task. There were 230 such tasks. Each model, except BEAST, was fitted to the aggregate choice rates of the 182 of these tasks that were part of CPC18's training data, using a grid search over the parameter space. BEAST was not fitted to this data. The values of its free parameters reflect the best fit to all five blocks of all 90 training problems from CPC15,[5] which are a subset of CPC18 training data. The models then all predicted the aggregate choice in the 48 remaining tasks that were part of CPC18's competition data. Finally, we used those predictions as a foresight feature in XGB algorithm with hyperparameters tuned according to CPC18's train set subset of decisions under risk tasks without feedback. As additional features (beyond the foresight feature), we used the set of objective features that define each choice task. In this exercise, BEAST was compared to two versions of Cumulative Prospect Theory,[2] to the Priority Heuristic,[56] and to the Decision by Sampling model.[57] In addition, we also compared it to an "ensemble" model that includes all five foresight features (i.e., the predictions of all five behavioral models were used as features in addition to objective features). The SI provides details on the implementation of the various models and detailed results.

**Choices13k**

***Data***

The Choices13k dataset was originally presented by Bourgin et al.,[17] and includes 13,006 binary choice tasks. Tasks were generated by the task generation algorithm used in CPC15,[5] and are therefore all members of the same space used in CPC18 that extends it. Hence, they can all be described by the set of objective features in Table S2. Specifically, each choice task includes two options marked "A" and "B", between which participants in an online experiment chose repeatedly across five trials. The data includes, for each choice task, the proportion of times in which participants chose Option B.

Participants in the experiment, Amazon Mechanical Turk users, were each presented with 20 choice tasks. On average, each task was faced by 16 participants. Participants were paid $0.75 plus a 10% bonus on their winnings from one randomly selected task, unless their payoff was negative, in which case the bonus was set to zero. As in Peterson et al.,[7] we removed from the dataset tasks in which one of the options was ambiguous and tasks in which participants did not receive any feedback, resulting with a dataset containing 9,831 risky choice tasks in which participants made five consecutive choices with full feedback after each choice. Additional details of this dataset can be found in Peterson et al.[7] and Bourgin et al.[17] Figure S3 provides a visual representation of the wide coverage of this dataset, particularly in comparison with the data of CPC18. Table S5 summarizes the main differences between Choices13k and CPC18.

***Benchmark models***

We compared BEAST-GB to models developed in Peterson et al.[7] that includes details of these models. In particular, we present the performance of BEAST-GB in comparison to the performance of two models from that study: Neural PT and Context-Dependent (CD). Neural PT is a neural network stochastic variant of prospect theory[3] in which the model searches the entire class of possible payoff and probability transformation functions assumed in prospect theory. Note that the search is not only over the space of parameters of the functions, but the functional forms themselves. In a sense, Neural PT reflects the version of prospect theory that best captures the data, and Peterson et al. show that it indeed predicts better than many other variations of prospect theory (including cumulative prospect theory). CD is the model that (after sufficient training) performed best in Peterson et al.'s analysis of this data. It is a fully unconstrained neural network that takes all information about both gambles as input and produces the choice rate as output. Because it is unconstrained, it effectively allows the network to learn subjective transformations of both outcomes and probabilities of the gambles, but in ways that are sensitive to the context of the other gamble. The performance of these benchmark models was taken directly from the analysis in Peterson et al.[7]

***Error evaluation***

To evaluate the models' error in Choices13k, we followed the original pipeline used by Peterson et al.[7] Specifically, we performed 50 iterations of the following process. First, we split the data to 90% train set and 10% test set (choosing 983 choice tasks randomly for the latter). Then, we trained the model on an increasing proportion of the train set, ranging between 1% of the train set (88 choice tasks) to 100% of it (8848 choice tasks). Next, we used the trained model to predict the held-out test set and computed its MSE. That is, for each proportion of the train set, we computed 50 MSEs on the test set. The reported results are the average of these 50 MSEs. To statistically compare the performance of different models, we used paired t-tests over the resulting MSEs (for 100% of the training data).

To derive the predictions of the model for further analysis, we performed five repetitions of a 10-fold cross-validation procedure, so that each task's prediction was based on the average of exactly five predictions of BEAST-GB, each derived when the algorithm is trained on (a different) set of 90% of the data.

### *Using BEAST-GB to explain behavior*

We probed the differences between the predictions of BEAST and those of BEAST-GB in an iterative process of Scientific Regret Minimization,[15] a process in which the theoretical model is critiqued with respect to a more predictive but less interpretable model. The idea underlying this process is that errors of the theoretical model can result both from it missing predictable patterns and from noise. Because BEAST-GB predicts almost all predictable variation, using it to critique BEAST is more effective than using the (noisy) data itself, and especially since BEAST-GB is a derivation of BEAST.

In each iteration, we sorted the tasks by descending order of the squared error between the two models' predictions. We then examined the tasks with the largest errors, trying to identify what features of behavior BEAST-GB captures, but BEAST does not. Upon identifying a pattern, we linearly corrected the predictions of BEAST so that they were closer to those of BEAST-GB and then moved to the next iteration. To avoid increasing BEAST's complexity and reducing its interpretability, most of these corrections were statistical: We only changed the predictions of BEAST after they were derived. However, we also found a possible mechanistic correction to BEAST (changing the model itself before deriving its new predictions) that does little to the model's complexity and interpretability. We then implemented this correction, trained the new version of the model on the CPC18 training data, and derived the trained model's predictions for all three datasets we use in this paper (see SI).

## HAB22

### *Data*

HAB22 includes data assembled by He et al.[6] from 15 different experimental contexts. The data from these different contexts was originally published in seven distinct papers by various researchers.[5,30,58–62] In each experimental context, participants made multiple one-shot choices between binary lotteries with up to two outcomes without feedback. Hence, the experimental task here was different than that used in CPC18 and Choices13k. Moreover, some choice tasks in this dataset are very different than the tasks in the other two datasets. Specifically, the difference between the EVs of the lotteries in some choice tasks here is especially large. For example, one task involved a choice between 500 with probability .4 vs. 50 with probability .8, EV difference of 160, and another task involved a choice between 500 with probability .8 and 100 for certain, EV difference of 300. In both tasks, most participants failed to maximize EV. Figure S3 shows a 2-d visualization of the similarities and differences between all choice tasks used in this paper and highlights that in HAB22 there is a cluster of choice tasks very different than the rest. Table S5 presents further details on this dataset and compares its main properties with those of the other datasets. In total, the HAB22 data includes 1565 choice tasks, although some of these are identical but were run in different experimental contexts and are thus treated as distinct.

Originally, He et al.[6] used four additional experimental contexts in their analyses. However, the data in these contexts is not usable for the purpose of our model comparisons.[37] In three contexts, there was an indexing error resulting in mismatches between the task IDs in the raw data and the original task IDs. This unfortunately has led to a mismatch between the parameters defining each task and the choice rate associated with it in the data. Consequently, the measured performance of the behavioral models that He et al. trained was distorted. In a fourth context, participants faced many of the same choice tasks more than once. As a result, the same exact task was often included both in the train and test set of the behavioral models. Hence, we could not properly compare BEAST-GB to the behavioral models in these four contexts and chose to exclude them.

***Benchmark models***

We compare BEAST-GB to all 53 behavioral models that He et al. investigated for the mixed gambles domain (Table S6). Models are diverse and include a range of different assumptions about human risky choice. Details of these models can be found in He et al.[6] Under He et al.'s inclusion criteria, all behavioral models had to include precise functional forms that have analytically specified likelihood functions. This allowed fitting of each model to each individual in each experimental context separately. Yet, this also excluded the model BEAST whose prediction is used as a feature in BEAST-GB. Hence, we also derived the predictions of BEAST, without retraining of its parameters, and present them for comparison. As an additional benchmark, we also trained behavioral-theory-free deep neural networks, and report on them in the SI.

***Evaluation method***

In their original investigation, He et al. fitted each of the behavioral models to each individual participant separately, using a subset of the choice tasks that the participant faced, and evaluated the fitted models based on their ability to predict the choices of the same individuals in the other (test) choice tasks. We consider this "known" individuals prediction task in the SI. BEAST-GB is a model for the prediction of new (unfamiliar) participants in new choice tasks (and the best prediction for new participants is the prediction of the mean choice behavior of the population). Thus, and to be consistent with the rest of the current study, we evaluated the models based on their ability to predict the choice rates of a new sample of participants from the population (i.e., participants that the model had no access to during training) in new choice tasks. Hence, we first split the participants in each of the 15 experimental contexts to five folds. We then repeatedly used data of four folds of participants for training and predicted the data of participants in the last fold. This was done in addition to using He et al.'s original segmentation of the choice tasks in that experimental context to 10 folds, using only choice tasks in nine of these folds for training and predicting behavior in the $10^{th}$ fold. That is, the train data included choices of 80% of the participants in 90% of the choice tasks, whereas the test data included choices of the other 20% of participants in the other 10% of the choice tasks of each experimental context.

Since He et al. derived individual participant predictions for each choice task in each benchmark behavioral model, we averaged these original individual predictions across the participants in the train set to derive a prediction for the aggregate out-of-sample choice rate in the test-set task. BEAST-GB was trained on the aggregated choice rates in the training data (i.e., unlike the benchmark behavioral models, BEAST-GB did not use individual participant data for its training). This process was repeated 50 times with different combinations of participants and tasks for the test set (i.e., we essentially performed a double cross validation procedure, on participants and on tasks). The reported results are the average of these 50 runs. To statistically compare the performance of different models, we used paired t-tests over the 50 resulting MSEs.

**Context generalization**

In the analyses of context generalization, we used the HAB22 data, with the addition of another experimental context ("Stewart15_1C_uniform") that we previously excluded because in that experiment many tasks were faced by the same participants more than once. Thus, when models were trained and tested within context, using this additional context introduces data leakage: The train and test data include choices of the same people in the same tasks. Under context generalization, however, models always predicted out of context and so there were no data leakage issues. Hence, here, we used 16 experimental contexts. Excluding this dataset does not qualitatively change any of the results.

Specifically, we repeatedly trained BEAST-GB on exactly 15 experimental contexts and then generated its predictions for the 16$^{th}$ context. Note that the model could not use the *dataset* feature here, as its values differed between training and testing. We report on the model's performance in this task of context generalization in two ways. First, we simply computed the MSE and completeness of the model in each of the 16 unseen datasets separately and report the average of these 16 MSEs and completeness scores.

The second evaluation we used relies on the fact that the exact same choice tasks (i.e. choice between the same two payoff distributions) were at times used in different experimental contexts in HAB22. Specifically, there are 1221 unique choice tasks in HAB22, and 384 of these were independently used in more than one experimental context: 338 tasks were used in two contexts, 33 were used in three contexts, 12 were used in four contexts, and one task was used in five contexts. Thus, there were 828 instances where a choice task from the test set (the "16$^{th}$ experimental context") also appeared in the train set (at least once). For each of these 828 instances, we computed the prediction errors of BEAST-GB, and we report on the MSE across all these instances.

In addition, we computed the prediction error of an a-parametric model that predicts, in each instance, the observed choice rate of the same task in the training data. This allowed us to evaluate the error of BEAST-GB relative to a very strong benchmark that assumes behavior in the same task is similar across experimental contexts. Note that the expected error of this benchmark is the sampling variance, and so a model whose average prediction error is smaller than the average sampling variance should be more accurate than this benchmark,

To statistically examine the difference between BEAST-GB and this strong benchmark, we used paired t-test for the prediction errors across all 828 instances. Finally, we generated for each of the benchmark behavioral models in HAB22, a prediction for each instance by averaging all the model's *training* predictions of that choice task in the train data (i.e., in the 15 experimental contexts available for training). A training prediction here is the model's "prediction" for a participant's choice in a task that was part of the training of the model when it was originally fitted to the data. Hence, these predictions use the entire training data to provide a prediction for out-of-sample behavior in the test experimental context.

**Data availability**

Raw data for CPC18, as well as processed data for analyses of the previously published datasets (Choices13k and HAB22) are publicly available at https://doi.org/10.17605/OSF.IO/VW2SU

**Code availability**

Code for all models and analyses reported in this study is publicly available at https://doi.org/10.17605/OSF.IO/VW2SU

**Acknowledgments**

OP thanks Or David Agassi for help in analysis of some of the curated data. IE acknowledges support from the Israel Science Foundation (grant no. 1821/12). MT has received funding from the European Research Council (ERC) under the European Union's Horizon 2020 research and innovation programme (grant agreement n 740435). The funders had no role in study design, data collection and analysis, decision to publish or preparation of the manuscript.

**Author contributions**

OP, RA, EE, MT, and IE organized CPC18 (OP, EE, and IE designed the experiments and collected the experimental data; IE developed the first baseline model; OP, RA and IE programmed the baseline models; OP and EE managed submissions). DB, JCP, DR, TLG, and SJR submitted the winning model for the first track of CPC18. ECC and JFC submitted the winning model for the second track of CPC18. OP performed all post-competition analyses, including analyses of Choices13k and HAB22. OP wrote the manuscript, and all authors commented on it.

**Competing interests**

The authors declare no competing interests.

## Extended Data Tables and Figures

**Table S1.**

Best submissions to CPC18 first track and baseline models.

| Rank | ID | Type | MSE (x100) | p-value | P(agree) | Short description |
|---|---|---|---|---|---|---|
| 1 | BP47 | Hybrid | 0.569[a] | | 0.930 | BEAST-GB: See text |
| 2 | MS63 | Behavioral | 0.589 | 0.395 | 0.910 | BEAST.sd modified to have different weights to the best estimate of the expected value and the outcome of the mental simulations. Weights depend on the availability of feedback and the possibility of a loss. Additional noise when predictions are extreme |
| 3 | MS03 | Behavioral | 0.605 | 0.328 | 0.897 | Same as MS63 with additional biases favoring dominant options whose dominance structure is clear and options avoiding losses (or with low probability for losses) under several conditions concerning differences in number of outcomes, minimal outcomes, maximal outcomes, EVs, and modes of the two options. |
| 4 | HK73 | Behavioral | 0.613 | 0.284 | 0.897 | BEAST.sd modified to have different weights to the best estimate of the expected value and the outcome of the mental simulations, as a personal trait, and a possible alternation in choice in non-feedback trials. |
| 5 | KH04 | Behavioral | 0.614 | 0.269 | 0.900 | BEAST.sd modified to have different weights to the best estimate of the expected value and the outcome of the mental simulations, as a personal trait. |
| 6 | KH75 | Hybrid | 0.621 | 0.189 | 0.910 | Ensemble of 12 models: 5 similar to BEAST, 6 similar to Psychological Forest (differing in foresight prediction) and one logistic regression model submitted to CPC15 |
| 7 | CJ25 | Hybrid | 0.640 | 0.140 | 0.873 | Ensemble of BEAST.sd and a random forest algorithm using several insights from Psychological Forest, several new insights (e.g., difference in expected regret) and |
| | | | | | | several foresights, including BEAST.sd and cumulative prospect theory. |
| 8 | LC33 | Behavioral | 0.648 | 0.183 | 0.883 | BEAST.sd modified to have different weights to the best estimate of the expected value and the outcome of the mental simulations. Weights depend on the availability of feedback. |
| 9 | SA88 | Hybrid | 0.668 | 0.091 | 0.893 | Psychological Forest modified to use foresight BEAST.sd instead of BEAST, and two other features: one marking how distant the mean aggregate (predicted) behavior is from 50%, the other marking the (predicted) over-time trend in decision makers' choice. |
| 10 | SA49 | Hybrid | 0.672 | 0.082 | 0.897 | Psychological Forest modified to use BEAST.sd instead of BEAST as foresight, and a feature marking how distant the mean aggregate (predicted) behavior is from 50%. |
| − | Psych. Forest | Hybrid | 0.681 | 0.024 | 0.917 | Baseline: See text |
| 11 | HB89 | Behavioral | 0.692 | 0.108 | 0.927 | BEAST.sd modified to replace the EV part of the model with a utility model accounting for dispersion and skewness. |
| 12 | RY01 | Behavioral | 0.706 | 0.050 | 0.873 | BEAST.sd modified to increase noise when predictions are extreme. |
| − | BEAST.sd | Behavioral | 0.708 | 0.049 | 0.883 | Baseline: See text |

*Note.* Only submissions providing predictions not statistically worse than BEAST-GB are presented. To test for differences is predictive performance, we used a bootstrap procedure with 10,000 resamples from the competition set tasks and computed the proportion of times each model predicted better than BEAST-GB. P-values represent this proportion. P(agree) is the probability that the model and the data agree on the modal choice. That is, it is the proportion of times that the model classifies correctly the majority choice.

[a] This MSE is for the numeric predictions as submitted by the winning team. Results in the main text refer to the competition's organizers' replication of the submitted model and are slightly better (100*MSE = 0.556)

**Table S2**

Features in BEAST-GB

| Category and Name | Descriptive Label | Description |
|---|---|---|
| Objective | | |
| *Ha* | High payoff A | $H_A$ - High payoff in Option A. When Option A has multiple outcomes, *Ha* is the EV of the lottery in Option A. |
| *pHa* | Prob high payoff A | $pH_A$ - Probability of *Ha*. |
| *La* | Low payoff A | $L_A$ - Low payoff in Option A. |
| *LotShapeA*[a] | Shape lottery A | $LotShape_A$ - Shape of the distribution of the lottery in Option A (“R-Skew”, “L-Skew” or “Symm”). When Option A does not have multiple outcomes, *LotShapeA* = “-” |
| *LotNumA* | No. lottery outcomes A | $LotNum_A$ - Number of outcomes in distribution of the lottery in Option A. When Option A does not have multiple outcomes, *LotNumA* = 1 |
| *Hb* | High payoff B | $H_B$ - High payoff in Option B. When Option B has multiple outcomes, *Hb* is the EV of the lottery in Option B. |
| *pHb* | Prob high payoff B | $pH_B$ - Probability of *Hb*. |
| *Lb* | Low payoff B | $L_B$ - Low payoff in Option B. |
| *LotShapeB* | Shape lottery B | $LotShape_B$ - Shape of the distribution of the lottery in Option B (“R-Skew”, “L-Skew” or “Symm”). When Option B does not have multiple outcomes, *LotShapeB* = “-” |
| *LotNumB*[a] | No. Outcomes Lottery B | $LotNum_B$ - Number of outcomes in distribution of the lottery in Option B. When Option B does not have multiple outcomes, *LotNumB* = 1 |
| *Amb* | Ambiguous task | Indicator for an ambiguous choice task (1 if True, 0 otherwise) |
| *Corr*[a] | Options Correlation | Sign of correlation between generated payoffs in the two options (-1, 0, or 1) |
| *block* | Block no. | The block number in repeated choice tasks (each block corresponds to 5 trials) |
| *Feedback* | Feedback block | Indicator for block with feedback (1 if True, 0 otherwise) |
| *Dataset*[a] | Exp. context | Dataset from which task is taken. |
| Naive | | |
| *diffEVs* | Δ EVs | Difference between the payoff EV of Option B and the payoff EV of Option A. |
| *diffSDs* | Δ Std Devs | Difference between the payoff standard deviation of Option B and the payoff standard deviation of Option A. |
| *diffMins*[b] | Δ Min payoffs | Difference between the minimal payoff of Option B and the minimal payoff of Option A |
| *diffMaxs* | Δ Max payoffs | Difference between the maximal payoff of Option B and the maximal payoff of Option A |
| Psychological | | |
| *diffBEV0* | Δ Best-EV-estimates (no Fb) | Difference between the “best estimate” of the EVs as per BEAST, prior to getting feedback. When the tasks are not ambiguous *diffBEV0* = *diffEVs* |
| *diffBEVfb* | Δ Best-EV-estimates (w/Fb) | Difference between the “best estimate” of the EVs as per BEAST, after getting first feedback. When the tasks are not ambiguous *diffBEVfb* = *diffEVs* |
| *pBbet_Unbiased1* | Δ Probs-better-pay (no Fb) | Difference between the probability that Option B provides better payoff than Option A and the probability that Option A provides better payoff than Option B, as estimated by BEAST before getting feedback |
| *pBbet_UnbiasedFB* | Δ Probs-better-pay (w/Fb) | Difference between the probability that Option B provides better payoff than Option A and the probability that Option A provides better payoff than Option B, as estimated by BEAST after getting feedback |
| *diffUV* | Δ Uniform-pay-EVs | Difference between the EV of Option B when all its outcomes are transformed to be equally likely and the EV of Option A when all its outcomes are transformed to be equally likely. |
| *pBbet_Uniform* | Δ Probs-better-uniform-pay | Difference between the probability that Option B provides better payoff than Option A and the probability that Option A provides better payoff than Option B, when both options are transformed so that their outcomes are equally likely |
| *RatioMin* | Ratio min payoffs | Ratio between the smaller and the higher minimal outcomes of the two options. When the minimal outcomes have different signs, *RatioMin* = 0 |
| *SignMax*[a] | Sign max payoff | The sign of the maximal possible payoff in the task (-1, 0, or 1) |
| *diffSignEV* | Δ Sign-pay-EVs | Difference between the EV of Option B when all its outcomes are sign transformed and the EV of Option A when all its outcomes are sign transformed. |
| *pBbet_Sign1* | Δ Probs-better-sign-pay (no Fb) | Difference between the probability that Option B provides better payoff than Option A and the probability that Option A provides better payoff than Option B, as estimated by BEAST before getting feedback and after all payoffs are sign transformed |
| *pBbet_SignFB* | Δ Probs-better-sign-pay (w/Fb) | Difference between the probability that Option B provides better payoff than Option A and the probability that Option A provides better payoff than Option B, as estimated by BEAST after getting feedback and after all payoffs are sign transformed. |
| *Dom*[a] | Dominant option | Trinary indicator for the option that stochastically dominates another (1 = B dominates A; -1 = A dominates B; 0 = neither option has dominance) |
| Foresight | | |
| *BEASTpred* | BEAST prediction | The quantitative point prediction of BEAST for the choice task (and block). Predictions are made using the model's original implementation and without training it to new data (i.e., using parameters as found in Erev et al.[5]) |

*Notes.* This is an exhaustive list of every feature used in this paper as part of BEAST-GB. When run on particular datasets, some features may be completely constant and others may be duplicates of other existing features, in which cases these features are removed before running of the algorithm.

[a] Categorical feature which is dummy coded before running of the algorithm

[b] *diffMins* belongs to both the naïve and the psychological feature categories.

**Table S3.**

Submissions not statistically worse than the winner in the second, individual decision makers, track

| Rank | ID | MSE (x100) | *p*-value | Short description |
|---|---|---|---|---|
| – | Naïve | 9.399 | 0.631 | Baseline: See text |
| 1 | CC31 | 9.405 | | Winner: See text |
| 2 | CL34 | 9.415 | 0.494 | For target agents whose behavior in the non-target games is more similar to a variant of BEAST (Submission LC33 from the first track), use as prediction the variant of BEAST. For other agents, predict like the naïve baseline. |
| – | FM | 9.630 | See note | Baseline: See text |
| 3 | EH51 | 9.706 | 0.214 | Logistic regression with the following predictors: prediction of the naïve baseline, dummy for higher-EV option, target individual maximization rate in non-target problems, output of a logistic transformation of the difference between options' EVs, and several interactions between these predictors and a dummy for a non-ambiguous problem. |
| 4 | CJ26 | 9.803 | 0.192 | Ensemble of the naïve baseline and a random forest algorithm as in Submission CJ25 from the first track. |
| 5 | EC02 | 9.973 | 0.083 | A type of tree-based regression (Cubist) using each dimension that described a problem, various averages based on those dimensions and subject information, as well as features calculated by BEAST.sd. |

Note. Only models providing predictions not statistically worse than the winner are presented. To test for differences is predictive performance, we used a bootstrap procedure with 10,000 resamples from the competition set tasks and computed the proportion of times each model predicted better than BEAST-GB. *P*-values represent this proportion. The predictions of the Factorization Machine baseline were lost (after the MSE has been documented) and we could not compute the *p*-value without them.

**Table S4.**

Hyperparameter values used in each implementation of the Extreme Gradient Boosting algorithm trained

| Model and dataset | Hyperparameter | | | | | | |
|---|---|---|---|---|---|---|---|
| | eta | gamma | max_ depth | Min_ child_ weight | subsample | Colsample_ bytree | nrounds |
| BEAST-GB, CPC18[a] | 0.011 | 0.012 | 3 | 1 | 0.508 | 0.9912 | 978 |
| XGB, CPC18 decisions under risk with foresight: | | | | | | | |
| BEAST | 0.02 | 0 | 12 | 1 | 0.45 | 1 | 600 |
| Stochastic CPT | 0.02 | 0 | 13 | 2 | 0.45 | 1 | 2100 |
| Deterministic CPT | 0.01 | 0 | 8 | 1 | 0.5 | 1 | 500 |
| Decision by Sampling | 0.02 | 0 | 5 | 1 | 0.35 | 1 | 3500 |
| Priority Heuristic | 0.03 | 0 | 6 | 4 | 0.6 | 0.9 | 1400 |
| Ensemble of foresights | 0.015 | 0 | 4 | 1 | 0.4 | 1 | 2700 |
| BEAST-GB, Choices13k | 0.01 | 0.04 | 6 | 3 | 0.55 | 0.4 | 1900 |
| BEAST-GB, HAB22 | 0.01 | 0.01 | 5 | 3 | 0.25 | 0.55 | 1800 |

*Note.* Names of hyperparameters as in documentation of function xgb.train of package xgboost[63] in R
[a] The winning submission also included tuning for *alpha* and *lambda* regularization parameters, equaling 0.043 and 2.905 respectively.

**Table S5.**
Comparison between datasets used in this paper.

| | Dataset | | |
|---|---|---|---|
| | CPC18 | Choices13k | HAB22 |
| Number of choice tasks | 270 | 9831 | 1565[a] |
| Choice task properties: | | | |
| Number of trials in each task | 25 | 5 | 1 |
| Feedback after each choice? | First five trials without feedback, then full feedback | Full feedback | None |
| Number of outcomes in each lottery | Up to 10. | Up to 10 in one lottery and up to 2 in the other. | Up to 2. |
| Ambiguity possible? | Yes | No | No |
| Number of tasks per participant | 30 | 20 | Varies between 46 and 150 (mostly consistent within experimental context) |
| Number of participants per choice task | At least 90 | 16 on average | Varies between 15 and 122 (mostly consistent within experimental context) |
| Location | Physical labs in the Technion and HUJI | Amazon Mechanical Turk | Physical labs in various locations (except Stewart15_1C_positive_skew, and Stewart15_1C_uniform which was online) |
| Population | Mostly undergraduate students | MTurk workers | Students (Erev17app, Rieskamp_Positive, Stewart15_1A_negative_skew, Stewart15_1A_positive_skew, Stewart15_2A_negative_skew, Stewart15_2A_positive_skew, Stewart15_2B_negative_skew, Stewart15_2b_positive_skew, Stewart16), or pools of experimental participants (Fiedler12_exp1, Fiedler12_exp2, Pachur17, Pachur18_e1_session1, Pachur18_e1_session2, Stewart15_1C_positive_skew, Stewart15_1C_uniform). |
| Payment method | Payoff in 1 randomly selected task | Fixed proportion (10%) of payoff in 1 randomly selected task, but with minimal payoff of 0. | Payoff in 1 randomly selected task (Erev17app, Fiedler12_exp1, Fiedler12_exp2), fixed proportion of 1 randomly selected task (Rieskamp_Positive, Pachur17, Pachur18_e1_session1, Pachur18_e1_session2), hypothetical (Stewart15_1C_positive_skew, Stewart15_1C_uniform), or contingent on performance but unclear from methods exactly how (Stewart15_1A_negative_skew, Stewart15_1A_positive_skew, Stewart15_2A_negative_skew, Stewart15_2A_positive_skew, Stewart15_2B_negative_skew, Stewart15_2b_positive_skew, Stewart16) |

*Note.* [a] When HAB22 is used for Context Generalization analyses, it includes 1665 tasks. The 100 additional tasks come from an experimental context (Stewart15_1C_uniform) which includes many tasks that subjects faced twice within a session and was removed for the analysis in which models were trained and predicted within contexts. Under Context Generalization, when models predict behavior in new contexts, repeated choices were pooled together.

**Table S6.**
Benchmark models in HAB22

| Abbreviated name | Full name | Source |
|---|---|---|
| EV | Expected value | |
| EUT | Expected utility | Bernoulli, 1738 |
| Mean Variance | Portfolio theory w/ variance | Markowitz, 1952 |
| SEU | Subjective expected utility | Savage, 1954 |
| SEM | Subjective expected money | Edwards, 1955 |
| Heuristic RELM | Relative expected loss minimization | Edwards, 1956 |
| Variance Skewness | Mean, variance and skewness | Coombs & Pruitt, 1960 |
| Below-mean Semivariance | Below-mean semivariance | Fishburn, 1977 |
| Certainty Equivalence | Certainty equivalence theory | Handa, 1977 |
| Mean Alpha-target | Alpha target model | Fishburn, 1977 |
| Mean Below-target | Below target model | Fishburn, 1977 |
| Mean StdDev | Portfolio theory w/ standard deviation | Fishburn, 1977 |
| Mean Target Semivariance | Below-target semivariance | Fishburn, 1977 |
| Odds-based SWU | Odds-based subjective weighted utility | Karmarkar, 1978 |
| PT-Edit1 | Prospect theory | Kahneman & Tversky, 1979 |
| Heuristic BTA | Better than average | Thorngate, 1980 |
| Heuristic CC | Consequence count | Thorngate, 1980 |
| Heuristic EP | Equiprobable | Thorngate, 1980 |
| Heuristic LEPE | Low expected payoff elimination | Thorngate, 1980 |
| Heuristic LL | Least likely | Thorngate, 1980 |
| Heuristic LPE | Low payoff elimination | Thorngate, 1980 |
| Heuristic Minimax | Minimax | Thorngate, 1980 |
| Heuristic Minimax Regret | Minimax regret | Thorngate, 1980 |
| Heuristic ML | Mostly likely | Thorngate, 1980 |
| Heuristic MPW | Most probable winner | Thorngate, 1980 |
| Regret-Bell | Regret theory with expected value evaluation | Bell, 1982 |
| Regret-LS | Regret theory with expected utility evaluation | Loomes & Sugden, 1982 |
| Disappointment-Bell | Disappointment theory w/o rescaling | Bell, 1985 |
| Disappointment-LS1 | Disappointment theory w/ expected value evaluation | Loomes & Sugden, 1986 |
| Disappointment-LS2 | Disappointment theory w/ expected utility evaluation | Loomes & Sugden, 1986 |
| Dual-1 | Dual theory w/ hyperbolic weighting | Yaari, 1987 |
| Dual-2 | Dual theory w/ quadratic weighting | Yaari, 1987 |
| Noisy Retrieval | Prospective reference theory | Viscusi, 1989 |
| Venture | Venture theory | Hogarth & Einhorn, 1990 |
| CPT-LBW | Cumulative prospect theory w/ Lattimore et al.'s weighting | Lattimore et al., 1992 |
| CPT-TK | Cumulative prospect theory | Tversky & Kahneman, 1992 |
| DFT | Decision field theory | Busemeyer & Townsend, 1993 |
| Heuristic Similarity-Leland | Similarity model | Leland, 1994 |
| RAM | Rank-affected multiplicative weighting | Birnbaum, 1997 |
| CPT-Prelec | Cumulative prospect theory w/ Prelec's weighting | Prelec, 1998 |

| Subjective Expected Pleasure | Subjective expected pleasure | Mellers et al., 1999 |
|---|---|---|
| CoefVar | Coefficient of variation | Weber, 2004 |
| PT-Edit2 | Prospect theory w/ Wu et al.'s editing | Wu et al., 2005 |
| Disappointment-DC1 | Generalized disappointment theory w/ expected value evaluation | Delquié & Cillo, 2006 |
| Disappointment-DC2 | Generalized disappointment theory w/ expected utility evaluation | Delquié & Cillo, 2006 |
| Heuristic Priority | Priority heuristic | Brandstatter et al., 2006 |
| Aspiration | Aspiration-level theory | Diecidue & van de Ven, 2008 |
| TAX | Transfer of attention exchange | Birnbaum, 2008 |
| Dual System-M | Dual systems w/ expected value evaluation | Mukherjee, 2010 |
| Salience | Salience theory | Bordalo et al., 2012 |
| Distracted DFT | Distracted decision field theory | Bhatia, 2014 |
| Dual System-LOB | Dual systems w/ expected utility evaluation | Loewenstein et al., 2015 |
| UWS | Utility-weighted sampling | Lieder et al., 2018 |

*Note.* Implementations of the models taken from He et al.[6]

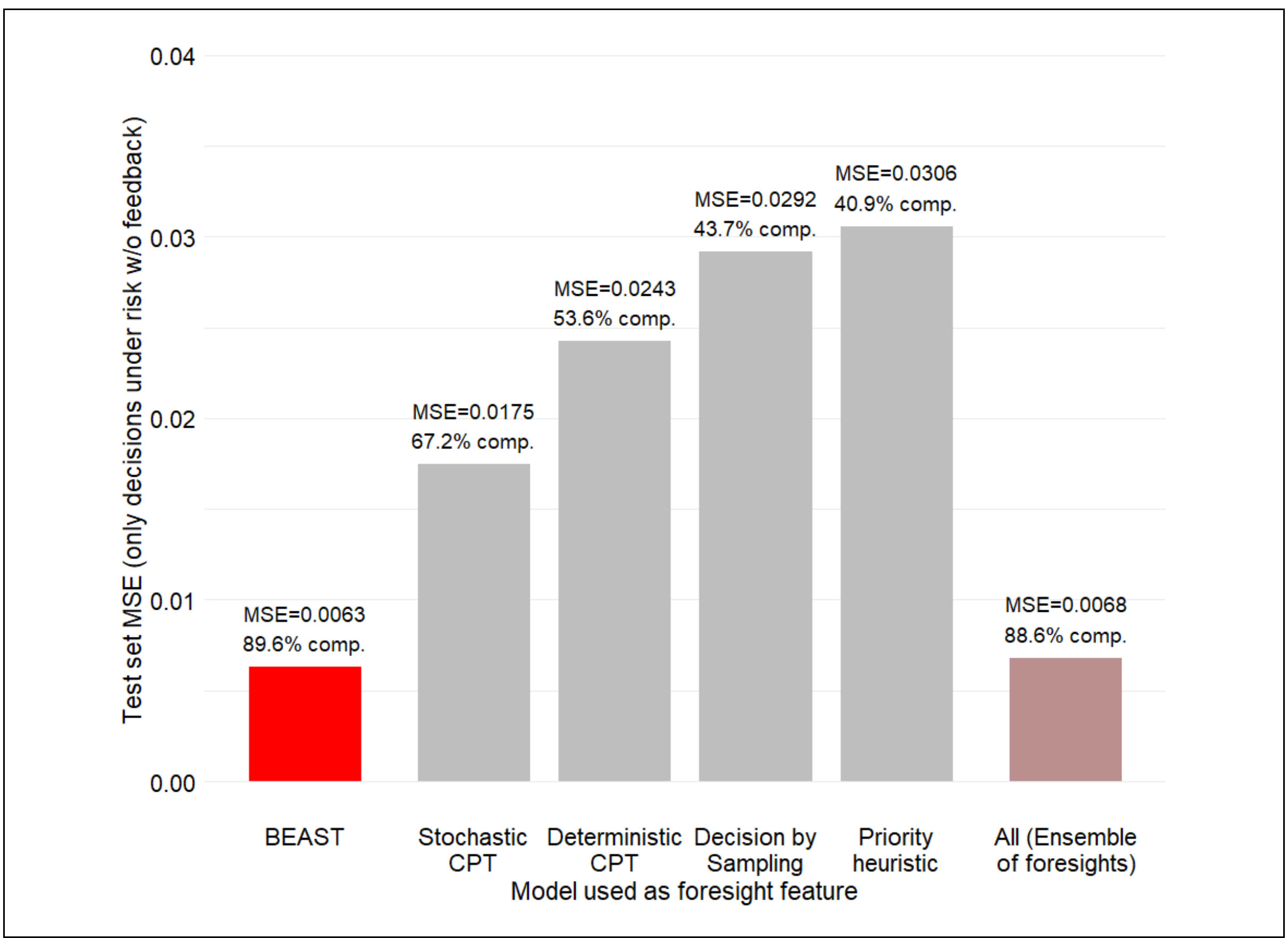

Figure S1. Comparison of the usefulness of behavioral models as foresight features. In each case, we tuned and trained an XGB algorithm using only the objective features (see Table S2) and the prediction of each foresight on CPC18's training data and predicted its test data. Both training and testing were restricted to the subset of CPC18's data that reflects pure decisions under risk (no feedback or ambiguity). All behavioral models except BEAST were first fitted to the training data independently to provide predictions. BEAST's predictions used the original parameters from CPC15.[5] Ensemble of foresights uses all five foresights combined.

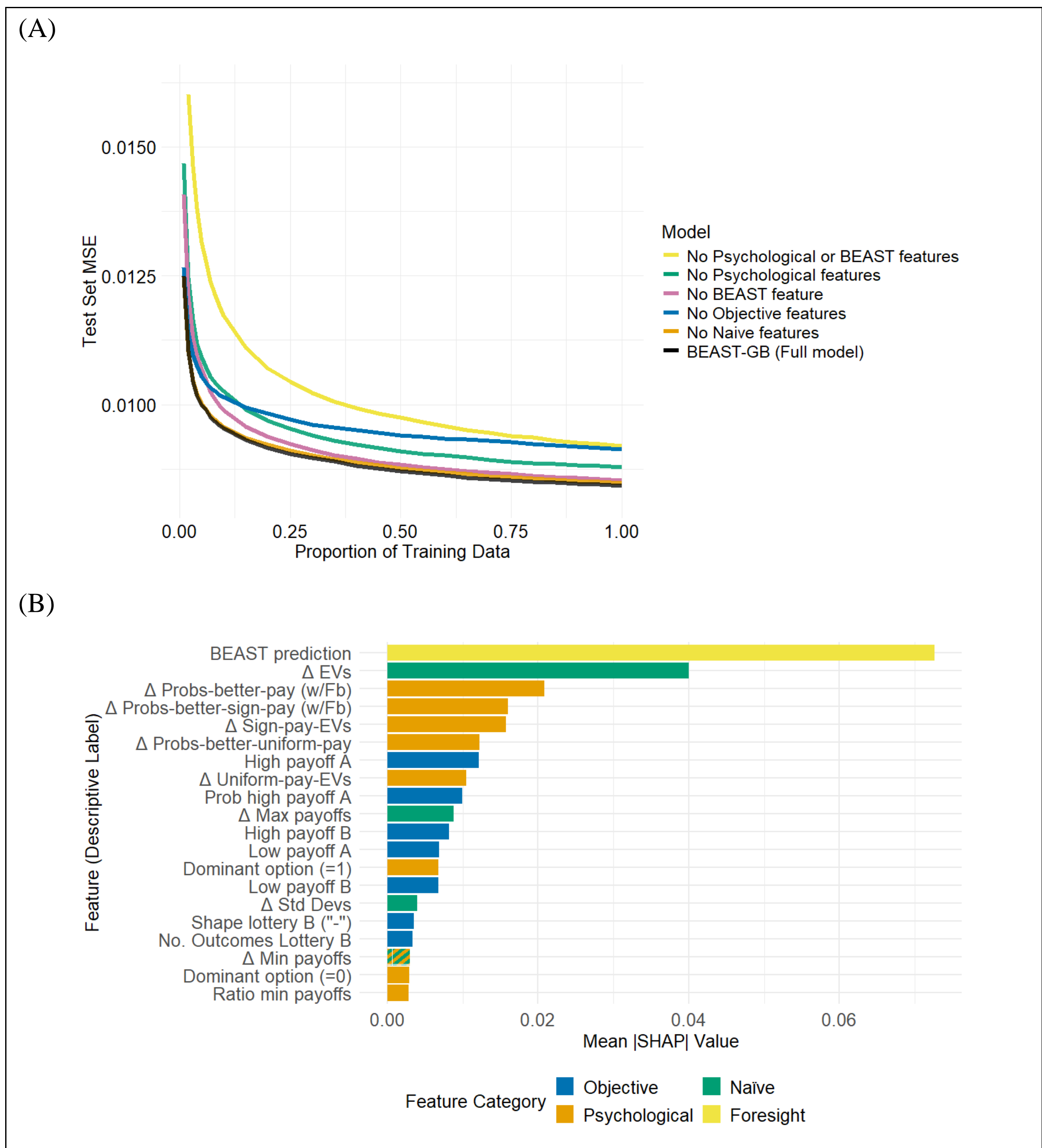

Figure S2. Feature importance analyses for Choices13k data. (A) Test set performance on Choices13k data when removing different sets of features from BEAST-GB. (B) Average absolute SHAP values of BEAST-GB's features in predicting Choices13k test data. Only top 20 features are shown. "Δ Min payoffs" is both a Naïve and a Psychological feature. Feature names and definitions in Table S2.

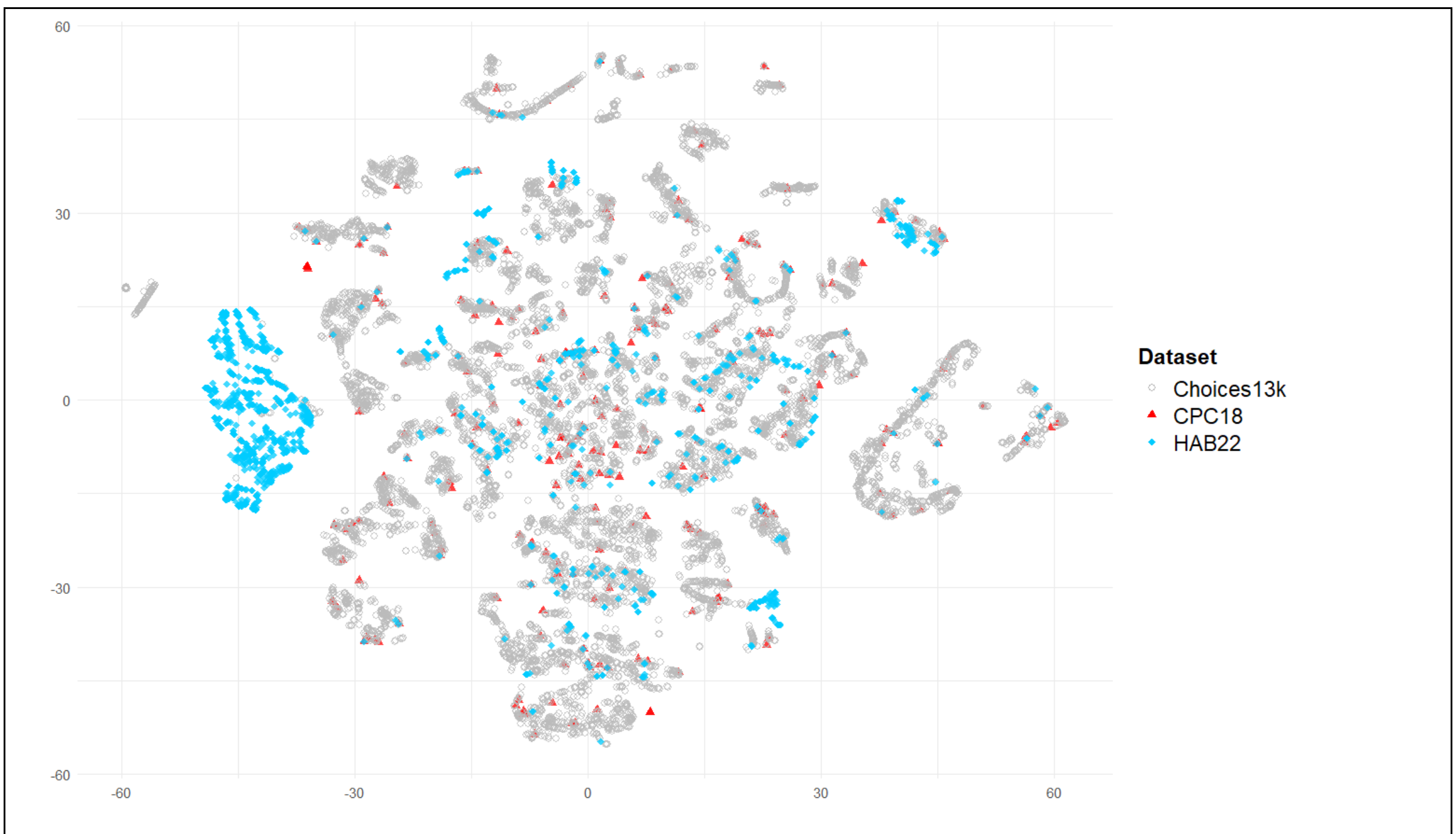

Figure S3. 2D visualization of all 11,666 choice tasks used in this paper. Each point is a single choice task represented in two dimensions obtained by implementing a t-SNE algorithm on the psychological feature space of the choice tasks. Tasks depicted closer together are conceptually more similar than tasks further apart. Choices13k data appears to cover well the space from which CPC18 data comes from, whereas HAB22 data is different than both.

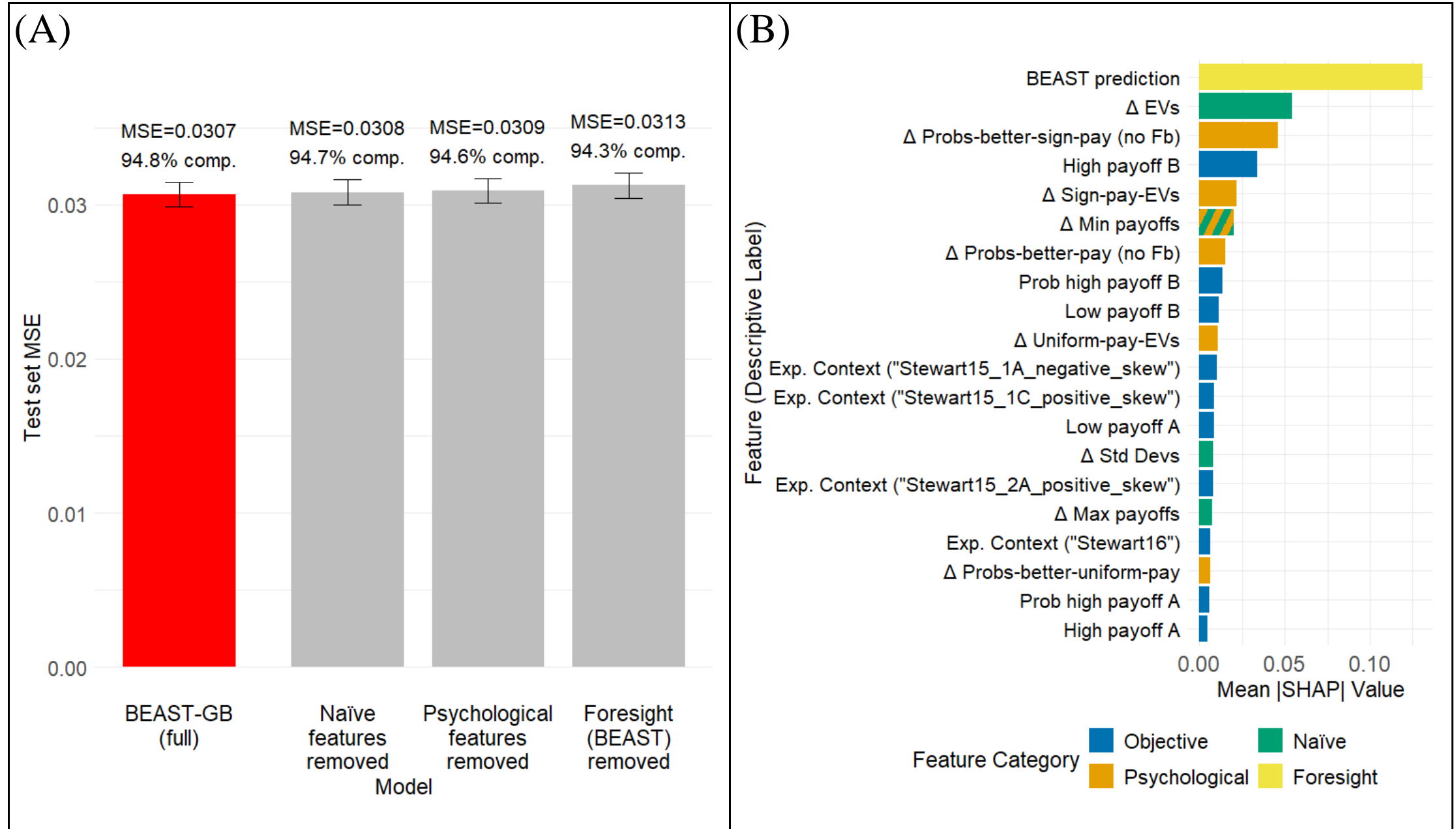

Figure S4. Feature importance analyses for HAB22 data. (A) HAB22 test set predictive performance of BEAST-GB and variations of it that remove different feature sets. Error bars represent ±1 SE for the mean across the 50 cross-validation iterations. (B) Average absolute SHAP values of BEAST-GB's features in predicting HAB22's test set. Only top 20 features are shown. "Δ Min payoffs" is both a Naïve and a Psychological feature. Feature names and definitions in Table S2.

## Supplemental Information:
## Predicting human decisions with behavioral theories and machine learning

### The model BEAST

BEAST (Best Estimate and Sampling Tools) is a model for decisions under risk, under ambiguity, and from experience[5]. It assumes that each option's payoff distribution is evaluated as the sum of three terms: the best estimate for that payoff distribution's expected value (EV), the average of a small sample of outcomes that are mentally drawn, and noise. Elements in the small mental sample are drawn using one of four sampling tools (that imply 4 behavioral mechanisms): Unbiased, Uniform, Sign, and Contingent Pessimism. Thus, BEAST relies in total on five main behavioral mechanisms. In addition, it assumes special handling of tasks that involve a stochastically dominant option. Hereafter, we provide the main properties of the model. For full details and implementation equations, we refer the reader to Erev et al., 2017.

Formally, the model assumes agent *i* chooses Option A over Option B after *r* trials with feedback if:

$$[BEV_A(r)_i - BEV_B(r)_i] + [ST_A(r)_i - ST_B(r)_i] + e(r)_i > 0$$

Where $[BEV_A(r)_i - BEV_B(r)_i]$ is the advantage of Option A over Option B based on the *Best Estimates* of their expected values; $[ST_A(r)_i - ST_B(r)_i]$ is the advantage of Option A over Option B based on mental sampling using *Sampling Tools*; and $e(r)_i$ is an error term. If a task is "trivial"—defined as one in which one option stochastically dominates the other—then $e(r)_i = 0$ for all *r*. Otherwise, this error term is normally distributed with mean 0 and standard deviation $\sigma_i>0$ (a property of agent *i*).[iv]

When an option is fully described, the "best estimate" of the EV (for all agents and in all trials) is simply the actual EV computed directly. When an option is ambiguous (does not include information on the probabilities of the possible payoffs), the best estimate of its EV is initially estimated using weighted average of (a) the EV of its alternative, (b) the EV computed under the assumption that all its outcomes are equally likely, and (c) its minimal outcome. The weight given to the minimal outcome, $0<\varphi_i<1$, is a measure of ambiguity aversion (a property of agent *i*). Each trial with feedback is then assumed to move the best estimate of the EV closer to the actual EV based on the observed payoff from that option.[v]

---

[iv] Hence, the inclusion of the feature *Dom* in the psychological insight features of BEAST-GB.

[v] In BEAST-GB, features *diffBEV0* and *diffBEVfb* capture the differences between the best estimates before and after getting feedback respectively.

The mental sampling process involves mentally sampling $\kappa_i$ (a property of agent *i*) outcomes from each option. Each sampling instance uses one of the four sampling tools, chosen independently of the other instances. In each instance, the same tool is used to sample both options. Sampling tool Unbiased is chosen with higher probability the more trials with feedback the agent sees, and $\theta_i>0$ captures agent *i*'s sensitivity to this feedback (the higher its value, the more likely it is that the Unbiased tool will be chosen, as a function of *r*). The likelihood of choosing any of the other sampling tools is equal and is contingent on $\beta_i>0$ which captures the magnitude of the agent's initial tendency to use one of the biased tools.

Sampling tool Unbiased implies a random unbiased draw. Before getting feedback, the draw is made from the options' described payoff distributions, but under the assumption that the options are positively correlated (a "luck-level" procedure, see Erev et al., 2017). When an option is ambiguous, the draw is made under the assumption that the minimal payoff is more likely than all other payoffs, and the other outcomes are equally likely. After getting feedback, the unbiased draw is always made from the history of observed outcomes. The use of this tool implies sensitivity to the probability that one option provides an outcome better than the other (which can also be described as sensitivity to the probability of regret).[vi]

Sampling tool Uniform implies a draw from a biased distribution that ignores all probability information. Rather, it assumes outcomes of an option are all equally likely (both before and after getting feedback). As the Unbiased tool, it also assumes the options are positively correlated.[vii]

Sampling tool Sign implies a draw from a biased distribution that focuses only on the payoff sign, ignoring outcome differences. Each payoff is assumed to be replaced with a constant that has the same sign as the original payoff, and then the draw is identical to the one made by the Unbiased tool.[viii]

Finally, sampling tool Contingent Pessimism implies "sampling" (with certainty) the minimal payoff of each option. Yet, this type of pessimism is triggered only if two conditions

---

[vi] In BEAST-GB, features *pBbet_Unbiased1* and *pBbet_UnbiasedFB* capture the differences between the probabilities of getting a better outcome when using the Unbiased tool, before and after getting feedback respectively.

[vii] In BEAST-GB, feature *pBbet_Uniform* captures the difference between the probabilities of getting a better outcome when using the Uniform tool. Feature *diffUV* captures the average expected difference between draws using the Uniform tool.

[viii] In BEAST-GB, features *pBbet_Sign1* and *pBbet_SignFB* capture the differences between the probabilities of getting a better outcome when using the Sign tool, before and after getting feedback respectively. Feature *diffSignEV* captures the average expected difference between draws using the Sign tool.

are met: the choice task includes at least one positive outcome, and the minimal outcomes appear dissimilar, with dissimilarity a function of their sign, their ratio, and the value of $0<\gamma_i<1$, a property of agent *i*. If either condition is not met, the tool is replaced by the sampling tool Uniform.[ix]

In total, BEAST includes six properties for each agent. The model, designed for prediction of the population-level choice rates in a task, assumes that these properties are drawn from uniform distributions defined as follows: $\sigma_i \sim U(0, \sigma)$, $\kappa_i \sim (1,2, 3, ..., \kappa)$, $\beta_i \sim U(0, \beta)$, $\theta_i \sim U(0, \theta)$, $\gamma_i \sim U(0, \gamma)$, and $\varphi_i \sim U(0, \varphi)$. The upper bounds of the distributions are the model's free parameters. In our analyses, we use the model with the parameter values obtained by Erev et al. (2017) when they fitted it on the train set of the 2015 choice prediction competition: $\sigma = 7$, $\kappa = 3$, $\beta = 2.6$, $\gamma = .5$, $\varphi = .07$, and $\theta = 1$.

## CPC18

### *Space of choice tasks*

Each choice task participants in CPC18 faced belongs to a 14-dimensional space of tasks. Two of the 14 dimensions, *Block* and *Feedback* vary within tasks. The 25 choice trials are divided into 5 blocks of 5 trials each, such that *Feedback* is absent in the 1st block and complete in the other four blocks. The other 12 dimensions uniquely define a choice task: 5 dimensions represent each payoff distribution, dimensions *Amb* sets whether Option B is ambiguous, and dimension *Corr* sets the correlation between the options' payoffs.

The 10 dimensions defining the payoff distributions of the options are: $L_A$, $H_A$, $pH_A$, $LotNum_A$, $LotShape_A$, $L_B$, $H_B$, $pH_B$, $LotNum_B$, $LotShape_B$. In particular, Option A provides a lottery, which has an expected value of $H_A$, with probability $pH_A$ and provides $L_A$ otherwise (with probability $1 - pH_A$). Similarly, Option B provides a lottery, which has an expected value of $H_B$, with probability $pH_B$, and provides $L_B$ otherwise (with probability $1 - pH_B$). The distribution of the lottery of Option A (Option B) around its expected value $H_A$ ($H_B$) is determined by the parameters $LotNum_A$ ($LotNum_B$) that defines the number of possible outcomes in the lottery, and $LotShape_A$ ($LotShape_B$) that defines whether the distribution around its mean is symmetric, right-skewed, left-skewed, or undefined (if $LotNum = 1$).

When a lottery is defined (i.e., $LotNum_A$ and/or $LotNum_B > 1$), its shape can be either "*Symm*", "*R-skew*," or "*L-skew*". When the shape equals "*Symm*" the lottery's possible

[ix] In BEAST-GB, feature *diffMins* captures the difference between the minimal outcomes, hence the draws made when using the Contingent Pessimism tool. Features *RatioMin* and *SignMax* capture the two conditions that trigger pessimism.

outcomes are generated by adding the following terms to its EV ($H_A$ or $H_B$): $-k/2$, $-k/2+1$, …, $k/2-1$, and $k/2$, where $k = LotNum - 1$ (hence the lottery—but not necessarily the option—has exactly *LotNum* possible outcomes). The lottery's distribution around its mean is binomial, with parameters $k$ and ½. In other words, the lottery's distribution is a form of discretization of a normal distribution with mean $H_A$ or $H_B$. Formally, if in a particular trial the lottery is drawn (which happens with probability $pH_A$ or $pH_B$), the outcome generated is:

$$\begin{cases} H - \frac{k}{2}, \text{with probability} \binom{k}{0}\left(\frac{1}{2}\right)^k \\ H - \frac{k}{2} + 1, \text{with probability} \binom{k}{1}\left(\frac{1}{2}\right)^k \\ \vdots \\ H - \frac{k}{2} + k, \text{with probability} \binom{k}{k}\left(\frac{1}{2}\right)^k \end{cases}$$

When the lottery's shape equals "*R-skew*," its possible outcomes are generated by adding the following terms to its EV: $C^+ + 2^1$, $C^+ + 2^2$, …, $C^+ + 2^n$, where $n = LotNum$ and $C^+ = -n - 1$. When the lottery's shape equals "*L-skew*," the possible outcomes are generated by adding the following terms to its EV: $C^- - 2^1$, $C^- - 2^2$, …, $C^- - 2^n$, where $C^- = n + 1$ (and $n = LotNum$). Note that $C^+$ and $C^-$ are constants that keep the lottery's distribution at either $H_A$ or $H_B$. In both cases (*R-skew* and *L-skew*), the lottery's distribution around its mean is a truncated geometric distribution with the parameter ½ (with the last term's probability adjusted up such that the distribution is well-defined). That is, the distribution is skewed: very large outcomes in *R-skew* and very small outcomes in *L-skew* are obtained with small probabilities.

For illustration, Figure 1 in the main text includes a choice between Option A that provides 50 with probability .2, 48 with probability .1, 44 with probability .1, and 1 otherwise (probability 0.6), and Option B that provides 16 with certainty. The values of the dimensions of this choice task are therefore as follows: $L_A = 1$, $H_A = 48$, $pH_A = 0.4$, $LotNum_A = 3$, $LotShape_A$ = "*L-skew*", $L_B = 16$, $H_B = 16$, $pH_B = 1$, $LotNum_B = 1$, $LotShape_B$ = "-". Here, in Option A, the lottery (which has expected value of 48, and is obtained with probability .4) includes three outcomes, thus $C^- = 4$ and the terms added to the EV are 2, 0, and −4. The other two dimensions that define this task are $Amb = 0$ (no ambiguity) and $Corr = 0$ (no correlation between the options). As mentioned above, *Block* and *Feedback* are studied within task.

***Task selection algorithm***

The 120 choice tasks in Experiments 1 and 2 were generated according to the following algorithm:

1. Draw randomly $EV_A$' ~ Uni(-10, 30) (a discrete uniform distribution)
2. Draw number of outcomes for Option A, $N_A$:
   2.1. With probability .4 ($N_A = 1$), set: $L_A = H_A = EV_A$'; $pH_A = 1$; $LotNum_A = 1$; and $LotShape_A$ = "-"
   2.2. With probability .6 ($N_A > 1$), draw $pH_A$ uniformly from the set {.01, .05, .1, .2, .25, .4, .5, .6, .75, .8, .9, .95, .99, 1}
      2.2.1. If $pH_A = 1$ then set $L_A = H_A = EV_A$'
      2.2.2. If $pH_A < 1$ then draw an outcome *temp* ~ Triangular[-50, $EV_A$', 120]
         2.2.2.1. If Round(*temp*) > $EV_A$' then set $H_A$ = Round(*temp*); $L_A = \text{Round}[(EV_A' - H_A \cdot pH_A)/(1 - pH_A)]$
         2.2.2.2. If Round(*temp*) < $EV_A$' then set $L_A$ = Round(*temp*); $H_A = \text{Round}\{[EV_A' - L_A(1 - pH_A)]/pH_A\}$
         2.2.2.3. If round(temp) = $EV_A$' then set $L_A = H_A = EV_A$'
      2.2.3. Set lottery for Option A:
         2.2.3.1. With probability 0.6 the lottery is degenerate. Set $LotNum_A = 1$ and $LotShape_A$ = "-"
         2.2.3.2. With probability 0.2 the lottery is skewed. Draw *temp* uniformly from the set {-7, -6, … ,-3, -2, 2, 3, … , 7, 8}
            2.2.3.2.1. If *temp* > 0 then set $LotNum_A$ = *temp* and $LotShape_A$ = "*R-skew*"
            2.2.3.2.2. If *temp* < 0 then set $LotNum_A$ = -*temp* and $LotShape_A$ = "*L-skew*"
         2.2.3.3. With probability 0.2 the lottery is symmetric. Set $LotShape_A$ = "Symm" and draw $LotNum_A$ uniformly from the set {3, 5, 7, 9}
3. Draw difference in expected values between options, *DEV*: $\boldsymbol{DEV} = \frac{1}{5}\sum_{i=1}^{5} U_i$, where $U_i$ ~ Uni[-20, 20]
4. Set $EV_B' = EV_A + DEV$, where $EV_A$ is the real expected value of Option A.
   4.1. If $EV_B' < -50$ stop and restart the process
5. Draw $pH_B$ uniformly from the set {.01, .05, .1, .2, .25, .4, .5, .6, .75, .8, .9, .95, .99, 1}
   5.1. If $pH_B = 1$ then set $L_B = H_B = \text{Round}(EV_B')$

5.2. If $pH_B < 1$ then draw an outcome *temp* ~ Triangular[-50, $EV_B$', 120]

5.2.1. If Round(*temp*) > $EV_B$' ten set $H_B$ = Round(*temp*);

$L_B = \text{Round}[(EV_B' - H_B \cdot pH_B)/(1 - pH_B)]$

5.2.2. If Round(*temp*) < $EV_B$' then set $L_B$ = Round(*temp*);

$H_B = \text{Round}\{[EV_B' - L_B(1 - pH_B)]/pH_B\}$

6. Set lottery for Option B:

6.1. With probability 0.5 the lottery is degenerate. Set $LotNum_B$ = 1 and $LotShape_B$ = "-"

6.2. With probability 0.25 the lottery is skewed. Draw *temp* uniformly from the set {-7, -6, … ,-3, -2, 2, 3, … , 7, 8}

6.2.1. If *temp* > 0 then set $LotNum_B$ = *temp* and $LotShape_B$ = "*R-skew*"

6.2.2. If *temp* < 0 then set $LotNum_B$ = -*temp* and $LotShape_B$ = "*L-skew*"

6.3. With probability 0.25 the lottery is symmetric. Set $LotShape_B$ = "Symm" and draw $LotNum_B$ uniformly from the set {3, 5, 7, 9}

7. Draw *Corr*: 0 with probability .8; 1 with probability .1; -1 with probability .1

8. Draw *Amb*: 0 with probability .8; 1 otherwise.

In addition, in the following cases the generated task is discarded for technical reasons: (a) there was a positive probability for an outcome larger than 256 or an outcome smaller than -50; (b) options were indistinguishable from participants' perspectives (i.e., had the same distributions and *Amb* = 0); (c) *Amb* = 1, but Option B had only one possible outcome; and (d) at least one option had no variance, but the options were correlated.

Moreover, tasks in Experiment 2 were selected using a stratified sampling procedure from a large pool of tasks selected according to the above algorithm. This procedure aimed to produce for Experiment 2 roughly the same number of tasks of the types "each option up to 2 outcomes", "exactly one option with more than 2 outcomes", and "both options with more than 2 outcomes" as their numbers in Experiment 1.

***Baseline models***

The organizers of CPC18 presented two baseline models, both heavily influenced by the model BEAST [5] detailed above. The first baseline model presented, *BEAST.sd* (BEAST subjective dominance), is a purely behavioral (i.e., includes no elements of statistical learning) extension of BEAST which changes the definition of a "trivial" choice task that has reduced noise. BEAST used an objective definition (the existence of stochastic dominance), whereas BEAST.sd uses a subjective definition. Specifically, a task is likely to be perceived as trivial if both the EV rule and the equal weighting rule favor the same prospect, and the choice of that prospect does not lead to immediate regret. BEAST.sd further assumes that in complex tasks (a task in which one option has at least 2 possible outcomes and the other has at least 3 possible outcomes), the estimation noise is increased. Finally, BEAST.sd assumes faster learning from feedback in ambiguous tasks.

The second baseline, psychological forest[16], is a hybrid model using a random forest algorithm with 31 features (see Table S2). Fourteen of the features, *La, Ha, pHa, LotNumA, LotShapeA*, *Lb, Hb, pHb, LotNumB, LotShapeB*, *Amb, Corr, Block,* and *Feedback* capture the corresponding dimensions that define the choice task (see section *Space of choice tasks*). Sixteen additional features are behavioral insights. Four of these were defined by the developers of psychological forest as “naïve”, as they represent basic domain knowledge likely to be integrated into an algorithm even without deep knowledge of behavioral theories. These include the difference between the payoff distribution’s expected values, the difference between their standard deviations, the difference between their minimal outcomes, and the difference between their maximal outcomes. Twelve additional features were considered “psychological”. They were hand-crafted in direct relation to the underlying logic of BEAST; each inspired by at least one behavioral mechanism in BEAST. For example, to capture sensitivity to the probability of regret, psychological forest includes the difference between the probability that Option A provides a better payoff than Option B and the probability that Option B provides a better payoff than option A. Positive (negative) values of this feature imply that Option A (B) is more likely to lead to less immediate regret than Option B (A). The mathematical equations defining the features all appear in the original psychological forest paper. Psychological Forest was originally created using package randomForest[64] in R, using the default set of hyperparameters for regression. In particular, at each split, one third of the features was considered for splitting the data. Open-source code for both baselines, in several programming languages, is available through the competition’s website (https://cpc-18.com).

***Individual decision makers prediction challenge***

CPC18 included two parallel and independent challenges. The paper focuses on the first challenge involving the task of predicting the average population response in a new decision task. Here, we briefly describe the second challenge that involved the task of predicting the choices made by individual decision makers in pre-defined choice tasks.

**The prediction task.** The goal in this second challenge was to predict, for each of 30 “target” individual decision makers, the progression over time (in 5 time-blocks of 5 trials each) of the mean choice rate of Option B in each of five “target” choice tasks. Specifically, the organizers randomly selected 30 of the 240 decision makers who participated in Experiment 1 to be the target decision makers. For each of them, the organizers then randomly selected five of the 30 tasks they faced and removed their data from the training data available to participants in the competition. The training data in this track thus included complete sequences of 25 choices each, made by each of the 30 target individuals in 25 different (non-target) choice tasks (taken from the same space of tasks), as well as data regarding behavior of other (non-target) decision makers in the five target tasks. Thus, models submitted to this track had to provide 750 predictions in the range [0, 1] (30 target decision makers X 5 target tasks per decision maker X 5 blocks of choices per problem).

**Competition protocol.** Protocol for participation in this second challenge was slightly different and simpler than for the first challenge (which is described in the Methods section). Specifically, participants here were not required to submit their codes to the organizers, only their numeric predictions. The reason for this change is that participants in the second challenge knew in advance the (anonymous) identity of the target decision makers on which they were tested and their corresponding target tasks (the nature of the test tasks in the first challenge was unknown at time of submission). Beyond this change, the protocols for the two challenges were similar.

**Baseline models**. The organizers presented two baseline models for this second challenge. The first, naïve baseline, predicts that each individual target decision maker, in each block of its individual target tasks, would behave the same as the average decision maker behaves in the same block of that task. The average decision maker's behavior is estimated as the mean aggregate behavior of all decision makers for which training data exists (there are at least 90 decision makers for each such task).

Surprisingly, the organizers found it difficult to significantly outperform this naïve baseline. Using many statistical learning techniques, and employing knowledge extracted from the psychological literature (e.g. based on BEAST), the best baseline that they could find was the use of a Factorization Machine (FM),[65] a predictor based on Support Vector Machines and factorization models, which is employed in collaborative filtering settings (i.e., settings in which the goal is to generate predictions regarding the tastes of particular users, for whom some data exists, using information on the tastes of many other users, as in the Netflix Challenge). Each observation supplied to the baseline implementation of the FM is composed of a long binary feature vector with only two non-zero elements that correspond to the active decision maker and the active block within an active task. The response is the observed choice rate of the active decision maker in the active block of the active task (first transformed to imply the maximization rate of the problem, and then after making the prediction transformed back to implying the choice rate of Option B). Therefore, the FM model did not explicitly use the knowledge that behavior across different blocks of the same problem is likely correlated.

**Submissions and results.** Twelve submissions were made before the deadline. None of the submissions (nor the FM baseline model) provided better predictions than the naïve baseline. In fact, the winning submission (made by two of the authors of the paper, ECC and JFC) was conceptually very similar to this naïve baseline. The primary difference was that the choice-task and block-wise average was calculated with a 10-fold cross-validation procedure. The training data supplied to contestants were split into ten sets of training, validation, and test data such that each test and validation dataset included data only from those choice tasks that were known to be in the held-out dataset. The prediction for a given test observation in fold *i* on choice task *j* in block *k* was the average choice made in the training data in fold *i* for task *j* in block *k*. MSE was calculated for each of the ten validation datasets, and the fold with the lowest MSE was identified. The training data from that best performing fold was used to make predictions on the held-out data following the same choice-task and block-wise average procedure. Four submissions did not provide statistically inferior predictions to those of the winner. Table S3 provides details on these submissions. Importantly, all top submissions heavily relied on the predictions of the naïve baseline.

***Post competition survey of registrants***

After results of CPC18 were published, the organizers sent co-authors of registered teams E-mail invitations to complete a short anonymous survey regarding their effort and perceptions. A total of 72 invitations were sent (out of 82 registered persons to either competition track; to register, a team had to supply an Email address of only the lead author and the organizers could not recover the addresses of 10 co-authors), and 36 researchers answered the survey. Nine people indicated they were only registered to the first track of CPC18, seven indicated they were only registered to the second track, and 20 indicated they were registered to both tracks.

Responders came from diverse backgrounds, ranging from computer science and artificial intelligence to cognitive or mathematical psychology. The most common primary research field reported was behavioral economics (25% of respondents). Respondents indicated having moderate to extensive coding experience (*M* = 3.42, *SD* = 1.21, on a scale of 1 to 5) and a moderate amount of experience modeling human behavior (M = 2.94; SD = 1.45).

Twenty-four of the 29 responders who were registered to the first track indicated they also tried working on a submission to that track (only 3 stated they did not try to work on a submission, and 2 did not answer the question), and 16 of them submitted a model by the deadline. Those who tried working on a submission to the first track stated they spent on average 66.5 hours (*SD* = 92.2) working on CPC18, and that they tried developing an average of 12.6 different models on the data (*SD* = 33.3). Sixteen of the 24 also stated they were able to develop a model that outperforms the baseline models.

Nineteen of the 27 responders who were registered to the second track indicated they also tried working on a submission to that track (6 stated they did not try to work on a submission, and 2 did not answer the question), and 10 of them submitted a model by the deadline. Those who tried working on a submission to the second track stated they spent on average 84.7 hours (*SD* = 118.9) working on CPC18, and that they tried developing an average of 29.1 different models to the data (*SD* = 59.6). Only 5 of the 19 stated they were able to develop a model that outperforms the baseline models in this track. Note the reported averages for hours spent on CPC18 and numbers of models developed for the two tracks include in some cases the same response (for persons working on submissions to both tracks), and thus they should not be interpreted as the mean effort invested in each track, but as a general effort invested in CPC18.

***Foresight comparisons implementation details***

**Cumulative prospect theory.** We compared BEAST to two versions of cumulative prospect theory (CPT)[2]: deterministic and stochastic. The only difference between the two versions was that in the stochastic version, CPT's weighted values of the options' prospects, *WV*(*A*) and *WV*(*B*), were transformed to a probabilistic prediction for choice of Option B over Option A using a standard logit transformation:

$$P\left(B \succ A\right) = \frac{e^{\mu WV(B)}}{e^{\mu WV(A)} + e^{\mu WV(B)}}$$

where μ captures the sensitivity to the difference between the weighted values.

For both versions, we used the following specification for CPT. The weighted value for a prospect with possible outcomes $x_1 \le \ldots \le x_k \le 0 \le x_{k+1} \le \ldots \le x_n$ is:

$$WV\left(X\right) = \sum_{i=1}^{k} \pi_i^- u\left(x_i\right) + \sum_{j=k+1}^{n} \pi_j^+ u\left(x_j\right)$$

where $u(x)=\begin{cases} x^{\alpha}, & if\ x\geq 0 \\ -\lambda(-x)^{\alpha}, & if\ x<0 \end{cases}$ is a subjective utility function with α a diminishing sensitivity parameter (note we use the same parameter for gains and losses, following the suggestion in[66], and λ a loss aversion parameter, and:

$$\pi_1^- = w(p_1)$$
$$\pi_n^+ = w(p_n)$$
$$\pi_i^- = w(p_1+\ldots+p_i)-w(p_1+\ldots+p_{i-1})$$
$$\pi_j^+ = w(p_j+\ldots+p_n)-w(p_{j+1}+\ldots+p_n)$$
$$w(p)=\frac{\delta p^{\gamma}}{\delta p^{\gamma}+(1-p)^{\gamma}}$$

The latter is a two-parameter subjective weighting function[67] with γ a probability sensitivity parameter and δ the function's elevation parameter.

For the deterministic version, best fit for the training data was obtained for α = 0.88, γ = 0.89, δ = 0.9, λ = 1.2. For the stochastic version, best fit was obtained for α = 0.91, γ = 0.84, δ = 0.83, λ = 1.14, and μ = 0.25. We derived the trained models' predictions for the 48 test set decisions under risk tasks in CPC18. The test MSE of the deterministic version is 0.1406. The test MSE of the stochastic version is 0.0198.

**Priority heuristic**. The priority heuristic (PH)[56] is said to be triggered only for choice between lotteries with similar EVs. Following others[68] we screened each problem according to the ratio between the options' EVs: If it is greater than 2, then PH is not triggered. We assumed instead that in such cases the option with the higher EV is selected. If the ratio is smaller than 2, PH is used as follows (note PH requires no fitting of parameters, so it was not technically fit to the training data):

1. Consider the minimal possible outcomes of the two options. If the difference between them exceeds an outcome aspiration level, stop and choose the option with the higher minimal outcome. Otherwise, continue to step 2. The outcome aspiration level is 1/10 of the highest absolute possible outcome in the problem, rounded to the nearest prominent number (1, 2, 5, 10, 20, 50, 100, 200, 500 etc.).
2. Consider the probabilities associated with each of the minimal outcomes of the two options. If the difference between them exceeds 0.1, stop and choose the option with the lower probability for a minimal outcome. Otherwise, continue to step 3.

3. Consider the maximal possible outcome of the two options. If the difference between them exceeds the outcome aspiration level, stop and choose the option with the higher maximal outcome. Otherwise, continue to step 4.
4. Consider the probabilities associated with each of the maximal outcomes of the two options. If the difference between them exceeds 0.1, stop and choose the option with the higher probability for a maximal outcome. Otherwise, predict indifference between the lotteries.

We derived the model's predictions for the 48 test set decisions under risk tasks in CPC18. The test MSE of is 0.1707.

**Decision by sampling.** The decision by sampling model for risky choice[57,69] states that choice is set by a series of ordinal comparisons between target attribute values and a comparison attribute value. An accumulator tallies the number of favorable comparisons to one of the options and when the tally hits a threshold, the option that won more comparisons is chosen. Target attribute values are chosen randomly at each time step. Comparison attribute values are also chosen randomly, though they can be chosen either from the alternative option or from long term memory.

We used the implementation from http://www.stewart.warwick.ac.uk/software/DbS/source_code.html. As "context" that is used for long term memory retrievals, we used the supplied csv files from the same source, providing "real world distributions of amounts and probabilities". Before running the model, amounts (from the 'real-world distribution') were converted from British Pounds to Israeli Shekels at an exchange rate of 4.5 shekel per pound. Three free parameters were fitted to the decisions under risk subsample from the training data: outcome threshold and probability threshold, which are the minimal amount and probability by which a target attribute value should exceed the comparison attribute value to be considered favorable, and a choice threshold, which is the number of comparisons an option needs to win in order to be chosen. Best fit was obtained for the values 1, 0.1, and 1, for outcome threshold, probability threshold, and choice threshold respectively. We derived the trained model's predictions for the 48 test set decisions under risk tasks in CPC18. The test MSE of is 0.0434.

**Extreme Gradient Boosting models.** After deriving the predictions of each of the five behavioral models (BEAST,[x] Decision by Sampling, Priority Heuristic, and two versions of CPT) for each of the choice under risk tasks in the CPC18 data, we trained five Extreme

[x] The MSE of the (untrained) BEAST for the 48 test set decisions under risk tasks in CPC18 is 0.0100.

Gradient Boosting algorithms on the subset of tasks that were part of the CPC18 train data, using the 12 objective features defining each task and the one foresight-type feature which was the prediction of each of the behavioral models. In addition, we trained an ensemble model with all five foresights as behavioral features. We tuned, using package *caret* [70] the hyperparameters of each model via a grid search and five repetitions of 5-fold cross-validation procedure over the training tasks. Table S4 shows the values of the hyperparameters found in this procedure and used to train the full model.

**Choices13k**

***Bias-variance analysis***

To analyze the sources of different performance of BEAST-GB and its variant that does not include access to BEAST's predictions as a foresight feature, we performed a bias-variance analysis. To do this, we randomly divided the Choices13k dataset to 90% train and 10% held-out test set and then made 30 repetitions of the following process. First, we randomly sampled from the full training set proportion *p* of samples. Second, we trained each of the two models on that chosen sample. Finally, we derived the trained models' predictions for the held-out test set. This has given us, for each model and each proportion *p*, 30 predictions, which were used to compute squared bias (the mean squared difference between the average prediction and the observed choice rate) and the variance (the mean difference between the 30 predictions and the average prediction).

The results in Table SI.1 show that the error of the models is dominated by their bias rather than the variance. Importantly, with small training set size (1%, corresponding to 88 tasks) the difference between the biases of the two models is considerably larger than the difference between their variances. This shows that the availability of BEAST as a feature helps BEAST-GB trained on small data to have a low bias and thus low error. Yet, as the sample size increases, both differences become very small, suggesting that the removal of BEAST foresight from the model has negligible impact on the performance when there is sufficient data to learn from.

Table SI.1. Bias-variance analysis, with and without foresight feature, of Choices13k,

| | $Bias^2$ | | | Variance | | |
|---|---|---|---|---|---|---|
| p | With BEAST | Without BEAST | Difference | With BEAST | Without BEAST | Difference |
| 0.01 | 0.010038 | 0.011048 | 0.001011 | 1.55E-03 | 1.94E-03 | 0.000390 |
| 0.1 | 0.008264 | 0.008506 | 0.000242 | 6.42E-04 | 7.30E-04 | 0.000088 |
| 1 | 0.008001 | 0.008026 | 0.000025 | 2.66E-05 | 2.97E-05 | 0.000003 |

***Using BEAST-GB to explain behavior***

**Scientific regret minimization.** We iteratively examined the tasks in which the deviations of BEAST from BEAST-GB were the largest. The clearest initial difference observed was that the predictions in BEAST were too extreme. For example, while the average absolute difference of BEAST from the midpoint of 50% in the entire dataset was 23 percentage points (pp), for the top 100 tasks with largest deviations from BEAST-GB, the average absolute difference was 41 pp. Meanwhile, the difference of BEAST-GB's predictions from the midpoint in these 100 tasks was just 12 pp. Moreover, while in the full data, 46% of the tasks included an option with multiple (more than two) outcomes, in the top 100 tasks with largest deviations from BEAST-GB, there were 72 tasks with multiple outcomes. Finally, while in the full data, there were 17% of the tasks that included stochastically dominant options, in the top 100 tasks with largest differences from BEAST-GB there were only 3 such tasks. Together, and following previous findings,[28] we concluded that BEAST-GB predicts much noisier behavior in Choices13k than BEAST, and the noise is likely larger for more "complex" tasks (tasks that include an option with multiple outcomes) and smaller in "trivial" tasks (tasks with a dominant option). Indeed, we fitted a linear regression to predict the differences (in the full dataset) between BEAST and BEAST-GB on the difference of BEAST's prediction from the midpoint of 50% interacting with dummies for a complex, a regular, and a trivial task. The results confirm a strong significant association between BEAST's prediction from the midpoint and its difference from BEAST-GB. The association is also significantly stronger for "complex" tasks and weaker for "trivial" tasks. The output also showed that these predictors explain 72% of the variance in the differences between the two models' predictions. We thus created a new variable that corrects the predictions of BEAST according to this linear model. This allowed us to explore the differences that remain between BEAST-GB's predictions and the predictions of BEAST, corrected for noisier behavior.

In this second iteration of the process, we noticed that 77 of the top 100 tasks with largest remaining differences from BEAST-GB were tasks in which one option provided a sure loss while the other was a riskier option that allowed for a possible gain (but also a larger potential loss). In contrast, in the full data, only 12% of the tasks were of this type. Indeed, the analysis showed that in almost every case in these tasks, BEAST-GB makes riskier predictions than BEAST. We added dummy variables that capture this type of tasks to the linear regression predicting the deviations of BEAST from BEAST-GB. The results confirmed that BEAST appears to overlook a 'gain-seeking' pattern of behavior in the Choices13k dataset, predicting less riskier choices in these tasks. The addition of these variables increased the $R^2$ of the linear regression to 0.81.

In the third iteration of the process, we discovered that in the top 100 tasks with largest remaining differences, there were 7 tasks without any positive payoffs, and the median value of the psychological insight feature *RatioMin* among the other 93 tasks was 0.4. In the full dataset, there were only 3% of tasks without any positive payoffs and in the remaining tasks the median value of the feature *RatioMin* was 0.2. As per BEAST, the Contingent Pessimism sampling tool evokes usage of pessimism if two conditions are met: there is at least one positive outcome, and RatioMin is sufficiently small. Otherwise, the Contingent Pessimism sampling tool is replaced with the Uniform sampling tool. This implies that in the list of tasks with the largest remaining deviations between BEAST and BEAST-GB, there is an over-representation for tasks in which BEAST does not use pessimism but replaces it with the Uniform tool instead. Further, we found that in the tasks with the largest deviations and that in which BEAST potentially replaces pessimism with the Uniform tool, the distributions of the psychological features that capture the Uniform tool (*pBbetter_Uniform, diffUV*) very much differ from their distributions in the wider dataset. We thus suspected that BEAST's assumption that the Contingent Pessimism tool is replaced with the Uniform tool when conditions of pessimism are not triggered hurts the performance of the model. We congruently added to the linear regression variables that interact a dummy that captures tasks in which pessimism may not be triggered with features that capture both Uniform and Sign sampling tools. The results confirmed our suspicions, and the $R^2$ of the multiple regression increased to 0.88.

Finally, we discovered that while 45% of the tasks in the dataset include an option with a safe outcome, in the top 100 tasks with largest remaining differences there were 59 such tasks, and while in the full dataset there were 46% of the tasks with multiple outcomes, in the top 100 tasks there were 70 such tasks. Moreover, while only 30% of the tasks included a discrepancy between the option that maximizes EV and the option that maximizes the probability to choose the better option, in the top 100 tasks there were 65 such tasks. Finally, while in the full data in only 5% of tasks one option had a higher minimal outcome while another favored the EV rule, such that the difference between the EVs was larger than the difference between the minimal outcomes, in the top 100 tasks there were 37 such tasks. Together, and following previous research,[29] we concluded that for different task structures, with task structure defined by the number of outcomes in each option, the different mechanisms of BEAST may have different weights. Indeed, adding to the linear model explaining differences between BEAST and BEAST-GB a set of interactions between task structure and features that capture the five BEAST mechanisms increased the $R^2$ to over 0.90.

The output of this linear model could also be further analyzed, especially for the differential effects of the five mechanisms of BEAST in each "class" of tasks (1 outcome vs. 2 outcomes; 1 outcome vs. more than 2 outcomes; 2 outcomes vs. 2 outcomes; and 2 outcomes vs. more than 2 outcomes). The analyses confirmed that under different task structures, the various mechanisms are differently associated with the deviations between the models. For example, in tasks involving a sure outcome vs. many outcomes, BEAST appeared to overpredict the reliance on the difference between expected values, and underpredict the importance of the Uniform sampling tool. Note that in such tasks, using a "equally likely" heuristic indeed very much simplifies the decision task. In contrast, in simple tasks that involve a sure outcome vs. a gamble of exactly two outcomes, BEAST overpredicts the use of Uniform, but underpredicts the use of pessimism. Finally, BEAST underpredicts the use of sampling tool Sign in all task types, but more so when neither of the two options includes a sure option.

**Correction to BEAST.** Following these analysis, it became apparent that BEAST under-relies on the Sign sampling tool, and over-relies on the Uniform sampling tool in tasks in which the conditions that trigger the Contingent Pessimism tool are not met. As mentioned above, when these conditions are not met, BEAST assumes Uniform is used instead of Contingent Pessimism. Thus, we decided to replace this assumption with the assumption that when Contingent Pessimism's conditions are not met, the Sign tool is used instead. Note this change does not increase the complexity or reduces interpretability of the model, while other potential changes like dynamically adjusting the usage of the different sampling tools according to task structure will almost surely increase the number of free parameters in the model and its complexity.

With this simple single change, we estimated BEAST's six free parameters on CPC18's training data. This change indeed improved the fit of the model. The best fit parameters were $\sigma = 14$, $\kappa = 3$, $\beta = 2.3$, $\gamma = .5$, $\varphi = .07$, and $\theta = 1$. We then derived the predictions of this corrected version of BEAST (with these parameter values fixed) for each of the three datasets we use in this paper. The results show that the corrected version predicts more accurately in each of the datasets. In CPC18, the test MSE was reduced from 0.0079 to 0.0077; in Choices13k, the test MSE was reduced from 0.0216 to 0.0194; and in HAB22 the test MSE was reduced from 0.0723 to 0.0560.

## HAB22

### *Neural network model*

We tried training a range of multi-layer perceptrons (deep neural networks) with different architectures and hyper-parameters and with objective features alone. All models we tried had much worse validation errors than the best models for this data. The network with the lowest validation error we could find included three hidden layers, with 128, 256, and 64 nodes respectively, and relu activation functions, as well as a linear output layer. Each hidden layer also had a l2 regularizer with 0.001 regularization factor and dropout of 0.2. We ran the model for 1000 epochs without early stopping, learning rate of 0.001, and batch size 256. The MSE of this model on the test set was 0.0571, placing it behind 28 of the models examined in this paper for this dataset.

### *Predicting in-sample participants' behavior*

In addition to the main analysis alluded to in the main text, we also evaluated how well each of the behavioral models and BEAST-GB predicts the mean aggregate choice rates of known (in-sample) participants in unknown (out-of-sample) tasks (see Footnote iii in the main text). That is, in both this and the main analysis, the models were trained on the same data, but their errors were evaluated on different test sets. Whereas in the main analysis, the goal was the prediction of the mean aggregate choice rates of a subset of participants that the model had no access to during training, in the current analysis, the goal was to predict the aggregate choice rates of the same participants the models were trained on, but in new tasks.

The results for all models are depicted in Figure SI.1. The MSE of BEAST-GB was 0.0086, whereas the second-best model, CPT-Prelec, had MSE of 0.0090. The difference between the models (tested using t-test on the paired differences of the 50 MSEs, one for each tasks X participants nested cross-validation fold) is significant: $t(49) = -2.09$, $p = 0.042$, $\Delta MSE = -0.0004$, 95%CI = [−0.0007, −0.0000].

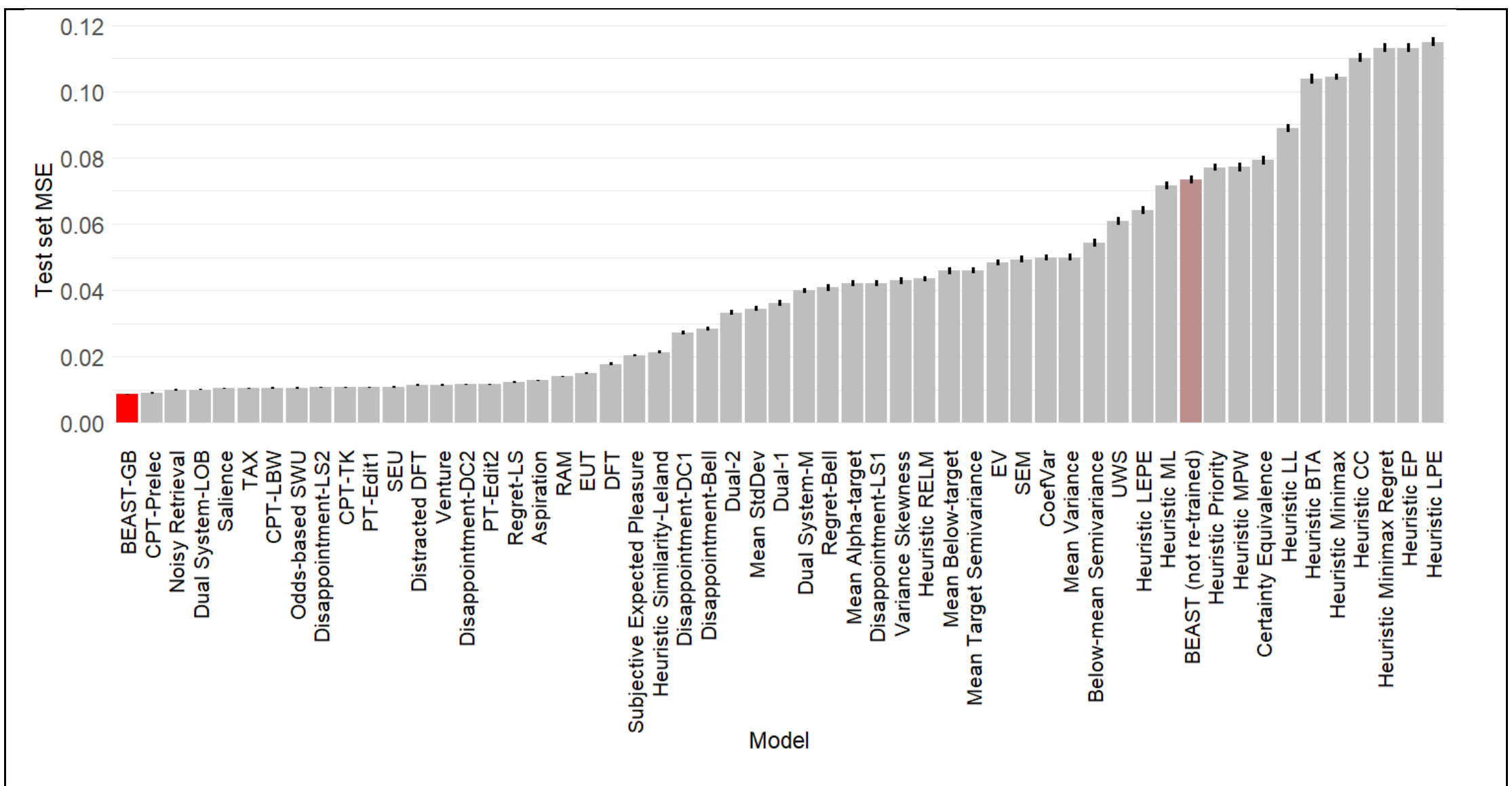

Figure SI.1 Test set performance on HAB22 data, for in-sample participants. In this analysis, models are evaluated in predicting the aggregate choice rates of known participants (80% of the sample) in new tasks (10% of the tasks). Error bars represent ±1 SE for the mean over the 50 test sets. Model names and sources in Table S6.

***Individual-level analyses***

Both BEAST and BEAST-GB are designed to predict population-level choice rates in decisions under risk and uncertainty, not choices of particular known individuals. Yet, the logic of BEAST, as implemented in BEAST-GB, as well as the prediction of BEAST-GB itself can help in development of designated individual-level choice prediction models. We demonstrated this in the HAB22 data which originally was used for comparison of behavioral models on the individual level and thus provided strong benchmarks. We employed two methods to develop these models, each relying to a different extent on the underlying logic of BEAST-GB.

First, we created, for each individual subject in each cross-validation (CV) iteration, a hybrid model very similar to BEAST-GB, but with two main differences. First, in addition to all the (task-level) features BEAST-GB already uses, we added one additional task-level feature: the population-level prediction of BEAST-GB itself in that task. Second, instead of using XGB as the algorithm that integrates all of the features together to derive a prediction, we used random forests.[24] Similarly to XGBs, random forests also combine many decision trees, each usually trained using a random subset of features and on a subset of the training sample. However, random forests use bagging rather than boosting: Instead of building the trees sequentially, each aiming to correct the prediction errors of the previous ensemble of trees, in random forests, trees are grown independently in parallel, and their predictions are then averaged. As a result, random forests are typically more robust to overfitting, especially in small datasets, utilize considerably fewer hyper-parameters (and are often robust to the chosen values of these), and run faster. While XGBs are often more accurate than random forests, because we train one model for each person in each CV iteration, a total of 6,580 models, and because the data available for individual-level training is rather small, random forests are more appropriate for the current setting than XGBs. Importantly, we did not tune the random forests' hyper-parameters to the data, but used the same default "out of the box" R *randomForest* implementation[64] for all models.

In our second method, we utilized a more traditional approach. We used the entire sample of participants while also modelling the individual differences in a hierarchical manner. Specifically, we implemented, in each of the 10 (task-level) cross-validation folds, a Bayesian mixed-effects logistic regression model that included fixed effects for the deviation of BEAST-GB from 0.5, and for five predictors that each capture one mechanism assumed by BEAST, and used as a feature in BEAST-GB. Specifically, the predictors are were difference between the EVs (feature *diffEV*, capturing sensitivity to expected payoffs; the "best estimate" mechanism in BEAST), the difference between the minimal outcomes (feature *diffMins*, capturing pessimism; the contingent pessimism sampling tool in BEAST), the difference between the probabilities that one option yields a better payoff than the other (feature *pBbetter_Unbiased1*, capturing sensitivity to immediate regret; the Unbiased sampling tool in BEAST), the difference between the probabilities that one option yields a better payoff sign than the other (feature *pBbetter_Sign1*, capturing sensitivity to payoff sign; the Sign sampling tool in BEAST), and the difference between the probabilities that one option provides a better outcome than the other had all the payoffs been equally likely (*pBbetter_Uniform,* capturing sensitivity to behave as if outcomes are equally weighted; the Uniform sampling tool in BEAST). Importantly, the model also included a random intercept and a random slope for each of the six predictors above. That is, the model aimed to capture the individual heterogeneity in subjects' sensitivity to the different mechanisms assumed by BEAST and to the level of similarity of the subject to the population-level predictions generated by BEAST-GB in each task. Models were estimated at the individual level, with the binary choices for each of the 658 subjects as the dependent variable. We used package brms [71] in R to fit the models. Each model included four chains with 1500 iterations, including 750 warm-up iterations, and default priors (flat prior over the reals for fixed effects, student-t(3, 0, 2.5) restricted to non-negatives prior for the random effects, and lkj(1) prior for the correlation matrix).

**Results.** Figure SI.2 presents the results of the individual-level predictive accuracy for the two methods we employed, in comparison with all other behavioral models used in HAB22, and for comparison, the accuracy of using the population-level predictions of BEAST-GB and BEAST (without re-training) as predictions for each individual. The MSE of the hybrid individual-level "behavioral random forests" models was 0.1552, lower than all behavioral models tested in the HAB22 data, except two, CPT-Prelec (MSE of 0.1537), and CPT-LBW (MSE of 0.1549). The MSE of the Bayesian mixed-effect logistic regressions was 0.1518, which is more accurate than all behavioral models, providing a new state-of-the-art for the individual-level predictions in this data. We used t-tests to compare the differences between the average errors of the models across subjects. The results showed that the errors of the behavioral random forests approach did not statistically differ than those of CPT-Prelec, $t(657) = -1.57$, $p = 0.118$, $\Delta MSE = -0.0015$, 95%CI = [−0.0034, 0.0004], or of CPT-LBW, $t(657) = -0.32$, $p = 0.746$, $\Delta MSE = -0.0003$, 95%CI = [−0.0023, 0.0017]. Yet, the mixed-effects logistic regression models, built using predictors heavily based on BEAST and BEAST-GB, were significantly more accurate for individual level data than CPT-Prelec, $t(657) = 2.24$, $p = 0.025$, $\Delta MSE = 0.0019$, 95%CI = [0.0002, 0.0036] and than CPT-LBW , $t(657) = 3.66$, $p < 0.001$, $\Delta MSE = 0.0031$, 95%CI = [0.0014, 0.0047].

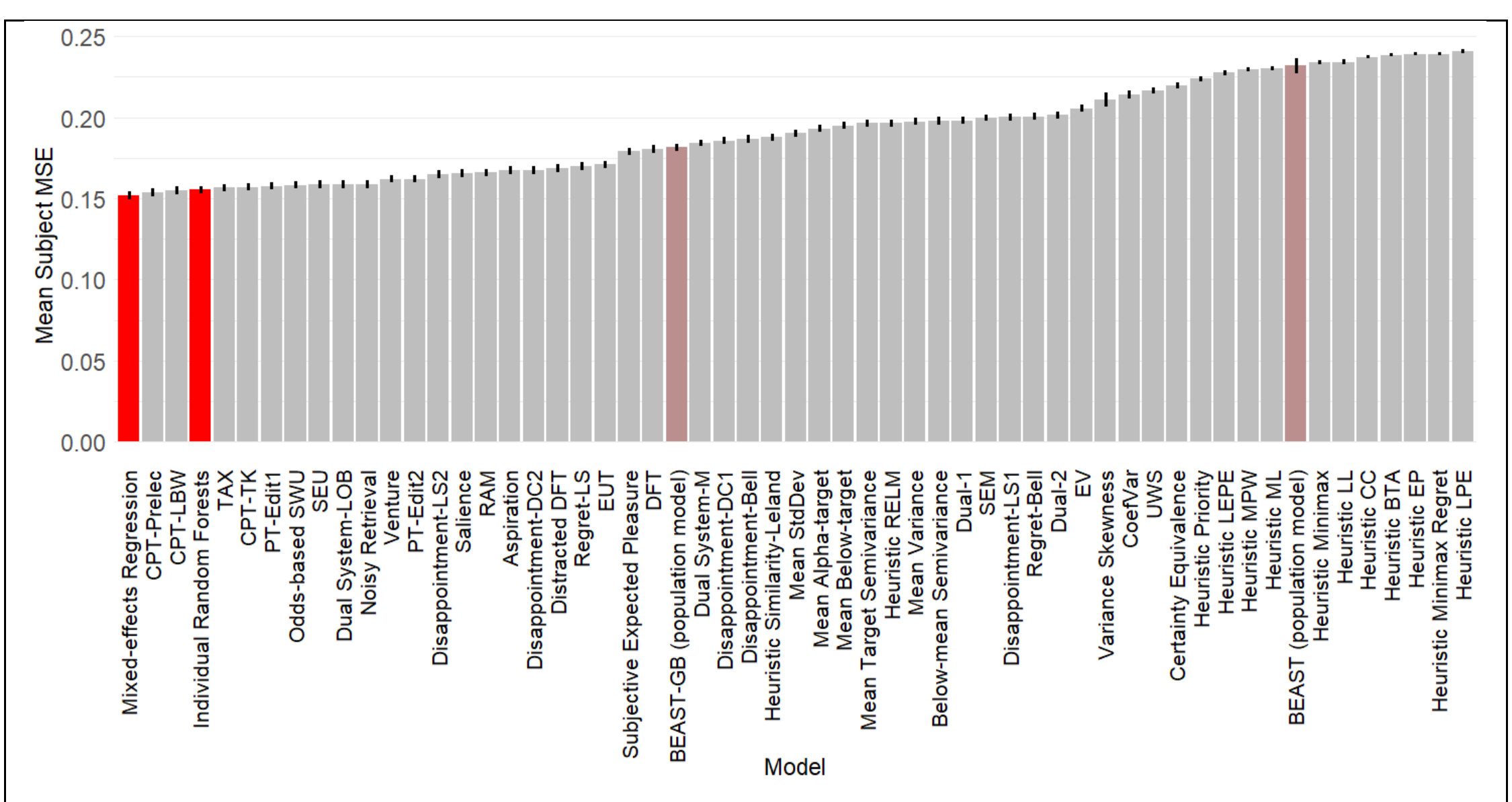

Figure SI.2. Predictive performance for individual-level binary data. Bars represent average MSE per subject in the HAB22 that involves data from 658 subjects. Error bars correspond to ±1 SE for mean. Mixed-effects regression and Individual Random Forests are the new individual-level models based on BEAST-GB that we developed. BEAST-GB and BEAST are population-level models on individual data (BEAST was not trained on any HAB22 data). Other models as in He et al., 2022.

We further analyzed the resulting coefficients of the mixed effects models. The average estimates of fixed effects, shown in Figure SI.3a, were stable across folds. In all models, the log-odds coefficient of the prediction of BEAST-GB was very large. On average it equaled 5.84, implying that when BEAST-GB changes from 0 to 1, the predicted probability changes from around 0.05 to 0.95, confirming that the population-level predictions of BEAST-GB are very strongly associated with the likelihoods of individual choices. Yet, the coefficients for all other predictors, except Uniform sampling (feature *pBbet_Uniform* in BEAST-GB), were also robustly estimated as different than zero, suggesting that on average participants were somewhat more sensitive to the difference between EVs, to the difference between the minimal outcomes (pessimism), and to Sign sampling, and were less sensitive to Unbiased sampling than BEAST-GB predicted.

Analyses of the random effects across predictors (Figure SI.3b) suggest that there was very little heterogeneity in pessimism and EV differences, but considerable heterogeneity in Sign sampling and adherence to the population-level prediction of BEAST-GB. Despite this large heterogeneity, nearly all participants' behaviors were positively associated with the predictions of BEAST-GB. Specifically, as these were mixed effect models, it was possible to compute the average total effect (fixed + random effect) of each predictor, for each subject. We computed the proportion of subjects who were estimated as “consistently sensitive” across folds to each of the predictors. Consistency was defined here as the average total effect having the same sign as the average effect minus two SDs over folds. This analysis shows that 98% of subjects had a consistently positive total effect for BEAST-GB, 74% had a consistently positive total effect for difference between EVs and 53% had a consistently positive total effect for the difference in minimal outcomes. None of the other predictors were consistent (either positively or negatively) for more than 50% of the subjects.

Finally, we examined whether the random slopes were related to the experimental context that the subjects faced. Figure SI.3c shows clear evidence that the estimated random effects tended to differ substantially between contexts. For example, in almost all experimental contexts by Stewart et al.,[30,60] the estimated effects of Sign sampling were considerably higher than in other contexts and the estimated effects of EV difference were smaller (indeed, Kruskal-Wallis tests confirm that the distributions in these estimates differed between contexts, $\chi^2(14) = 69.3$, $p < .001$ and $\chi^2(14) = 75.8$, $p < .001$, respectively). Yet, this was not true for all predictors. The Unbiased Sampling estimates did not differ by context ($\chi^2(14) = 20.6$, $p = .113$).

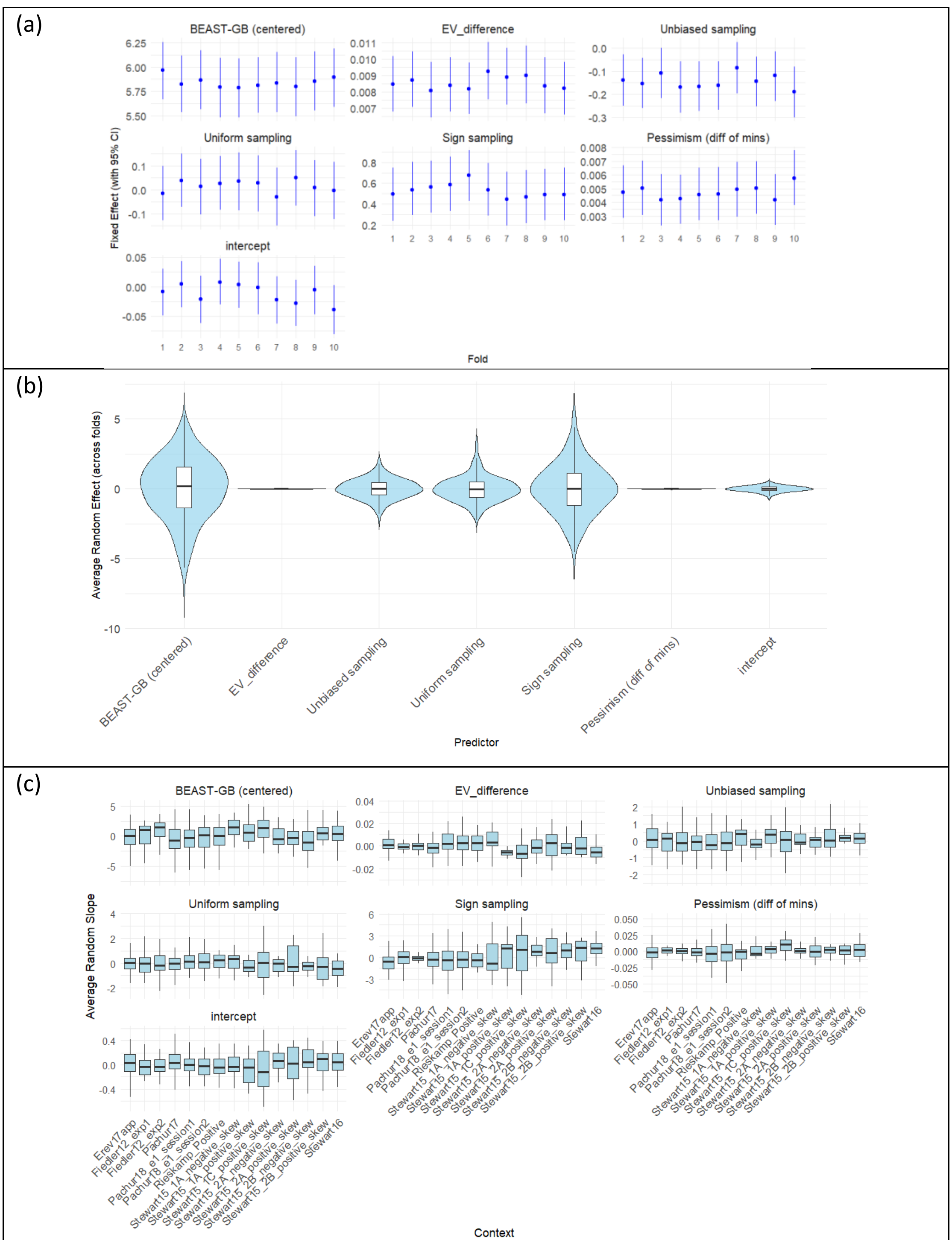

Figure SI.3. Analysis of Mixed-effects logistic regression outputs. (a) Comparison of fixed-effects estimates for each predictor, across folds. (b) Distributions of subject-level average random effects for each predictor (on average across 10 CV folds). (c) Subject-level random effects by predictor and context (on average across 10 CV folds). Context names as in He et al., 2022.

***Explaining prediction differences between BEAST-GB and BEAST***

For each of the 1565 tasks in HAB22's main analysis, we obtained five BEAST-GB predictions and five choice rates, one for each iteration of the cross-validation procedure over subjects. We analyzed this dataset of 7,825 predictions for the largest differences between the predictions of BEAST-GB and BEAST (which was not refitted to this data). We first found large differences across experimental contexts, such that in some contexts BEAST is heavily biased, predicting extreme choice rates. We fitted a linear regression to predict the deviations of BEAST from BEAST-GB's predictions using the interaction of experimental context with the difference of BEAST's prediction from the midpoint of 50%. The results showed that BEAST's predictions were too extreme in every context, but in some contexts the predictions were significantly more extreme. The $R^2$ of this simple model was 0.67, showing that simple linear correction of BEAST by context can go a long way. We then added to this linear model also interactions between features representing BEAST's five mechanisms (*diffEV*, *diffMins*, *pBbetter_Unbiased1*, *pBbetter_Sign1*, and *pBbetter_Uniform*, which represent sensitivity to EVs, and simulation tools contingent pessimism, unbiased, sign, and uniform respectively) with dummies capturing task structure (whether an option is a sure loss and whether an option is a sure gain) and the experimental context. The resulting model had $R^2$=0.81. This suggests that allowing for flexible adjustment of BEAST's mechanisms by task structure and experimental context accounts for a large portion of the deviation of BEAST (that is heavily biased in this data when not adjusted) from BEAST-GB (that predicts this data best).

**Additional datasets: Extensive form games**

We checked for the robustness of our proposed method in two additional datasets of human decisions in extensive form games [72]. To avoid having many researchers' degrees of freedom [73] (i.e. being able to select the method and/or analyses contingent on the outputs of the analyses), we repeated the same procedure used in the development of BEAST-GB: We started with a descriptive model that was proposed as a baseline in a choice prediction competition (i.e. developed before the collection of the test data), decomposed it to its theoretical insights and then used both the predictions of the model and its psychological insights as features within a XGB algorithm, in addition to the objective features defining each task.

***Data***

The data comes from two related choice prediction competitions for human decisions in simple extensive form games[72]. It includes 240 two-person games in which each of the players simultaneously chose between one of two actions. Player 1 (P1) chose either *in* or *out*, and Player 2 (P2) chose either *right* or *left*. If P1 chose *out*, the choice of P2 does not impact the payoff of either player, but if P1 chose *in,* the two payoffs are dictated by the choice of P2. The games, uniquely defined by 6 game-parameters, include versions of the ultimatum game,[74] the dictator game,[75] the trust game,[76] and the gift exchange game.[77] Games were explicitly described to both players, and there was no feedback nor communication between the players. The experiments used the strategy method, whereby players marked their choices without knowledge regarding the choices the other players made. Therefore, in each pair, choices made by the two players were independent. This enabled two independent competitions: The first competition was for the prediction of the choices of P1s, and the second competition was for the predictions of the choices of P2s. In each competition, data regarding choice rates in 120 games (training data) was made public. The goal was to predict the choice rates in the other 120 games (test data).

***Models***

To facilitate development of models, the organizers of the two competitions presented their best baseline models trained on the train data. Baselines for both competitions were similar and assumed players considered one of seven strategies: (a) choosing rationally (according to the rational model), (b) choosing rationally, but in case of indifference choosing such that the other player's payoff is maximized, (c) maximize the worst personal payoff, (d) choose rationally, but assuming the other player chooses randomly (level-1), (e) maximize the joint payoff, (f) minimize the difference between the two players' payoffs, and (g) maximize the payoff of the player with the lowest payoff. The difference between the two baseline models was in the implied behavior in each strategy and in the parameters used. Moreover, not all seven strategies applied to both players (e.g. for the second player, choosing rationally and maximizing the worst payoff are the same). Each competition received 14 submissions. Thus, we compared our new hybrid models to 15 models designed to predict the choice rates.

To develop the hybrid models, we employed the same approach used in development of BEAST-GB. In each competition, we started with the baseline models and decomposed them to the theoretical mechanisms that they imply drive choice. Specifically, the behavioral insights we derived were the predictions made by each of the seven strategies assumed by the baseline models. In addition, we also used as foresight the predictions made by the baseline models themselves (fitted on the training data by the competitions' organizers). Finally, we also added the six game-parameters as "objective" features. We then trained a XGB algorithm with all these features to the training data (including a process of hyper-parameter tuning), and generated the predictions for the test data.

***Results***

The original baseline model for prediction of P1s was outperformed by seven models submitted to the competition with MSE = 0.00853. The winner of that competition obtained MSE = 0.00735. The new hybrid model that we created obtained MSE = 0.00652. Our model thus provides a new state-of-the-art for these data, outperforming the winner of the competition by 11%.

The baseline model for prediction of P2s was outperformed by 11 models submitted to the competition with MSE = 0.00415. The winner of that competition obtained MSE = 0.00346. The new hybrid model obtained MSE = 0.00393. Such a submission to the competition would have been ranked 5th. Analyses of the four submissions that obtained better predictions suggested that all of them used behavioral ideas that were not implemented in the current model (or the baseline). Therefore, the hybrid improved upon the baseline but did not outperform models that used additional behavioral elements it was not supplied with, particularly with so little training data to learn from.